\pdfoutput=1
\documentclass{article}
\usepackage{subcaption}
\usepackage{mathrsfs}
\usepackage{amsmath}
\usepackage{amsfonts}
\usepackage{graphicx}
\usepackage[hidelinks]{hyperref}
\usepackage{float}
\usepackage{iclr2027_conference,times}
\usepackage{amsmath,amsfonts,bm}

\def\eqref#1{equation~\ref{#1}}
\def\1{\bm{1}}

\DeclareMathAlphabet{\mathsfit}{\encodingdefault}{\sfdefault}{m}{sl}
\SetMathAlphabet{\mathsfit}{bold}{\encodingdefault}{\sfdefault}{bx}{n}

\usepackage{mathtools}
\newcommand{\defeq}{\coloneqq}
\usepackage{enumitem}
\setlist{topsep=2pt,itemsep=2pt,parsep=0pt,partopsep=0pt}

\usepackage{algorithm}
\usepackage{algorithmic}

\title{Factorization Regret mediates compositional generalization in latent space}

\author{John Schwarcz \\
Edmond \& Lily Safra Center for Brain Sciences (ELSC) \\
The Hebrew University of Jerusalem
}
\date{\today}

\iclrfinalcopy

\begin{document}
\maketitle
\lhead{Preprint} % running header; use \lhead{} for none

\begin{abstract}
Are there still barriers to generalization once all relevant variables are known? We consider the challenge of generalizing to a novel combination of task-relevant latent variables in the \emph{Cognitive Gridworld}, a stationary Partially Observable Markov Decision Process where observations are generated by latent variables with parametric interactions. We establish \emph{Factorization Regret} as an information-theoretic quantity measuring the contribution of latent variable interactions to task performance. Using this metric, we explain the sub-optimality of Echo State Networks and uncover a failure mode whereby confidence becomes decoupled from accuracy. This suggests that utilizing interactions between relevant variables is a non-trivial capability. We next introduce \emph{Representation Classification Chains} to show that latent variable interactions and how to infer the variables' values can be learned together through Reinforcement Learning. Solving this variational inference problem enables generalization to novel combinations of relevant variables. Lastly, we demonstrate that once interactions are learned, imagined experience is sufficient for zero-shot control. In summary, we develop and motivate a minimal, theoretically grounded setting for research into generalization from experience.
\end{abstract}

\section{Introduction}

As Machine Learning continues to solve increasingly challenging problems, unreliability remains a significant barrier. One central obstacle is that what \textit{can} be learned depends crucially on what \textit{has} been learned. For instance, inferring \textit{time of day} from \textit{ambient light} becomes much more reliable after learning about \textit{seasons}. A combinatorial explosion can emerge from the interactions between variables, but typically only a few are relevant to a given task. Thus, the world may be compressed into something much more manageable, insofar as one's goal is concerned \citep{Tishby2000TheIB, Roy2011FindingAP}. Yet, despite needing only to learn that which is immediately relevant for survival, animals develop an intuition about the natural world that is far more reliable than reasoning models \citep{Tenenbaum2018BuildingMT, SB}.

To explain this intuition, it helps to differentiate \textit{meaningful} intelligence from that which is trivial. \cite{chollet2019measure} defines intelligence as `skill-acquisition efficiency', emphasizing the capacity to build upon prior knowledge. This relies on compositional generalization, the capacity to generalize to an unseen combination of elements. In this paper, we consider compositional generalization with latent variables as the elemental units. The mutual information among three or more random variables is known as \textit{Interaction Information} \citep{Watanabe1960InformationTA}. Interaction Information has been used to understand sensory modality integration \citep{Proca2022SynergisticIS} and causal inference \citep{Ghassami2017InteractionIF}. In this paper we connect Interaction Information with the notion of \textit{Semantic Information} as the subset of information meaningful for survival \citep{Kolchinsky2018SemanticIA}. To exemplify their relationship, consider an agent with two actions (i.e., \textit{causal interventions}), where reward is determined by an \textit{exclusive or} (XOR) between the selected action and a binary variable. XOR demonstrates Interaction Information in its purest form \citep{Rosas2019QuantifyingHI}: Action and reward are independent unless the agent knows the value of the binary variable, so information about its value is semantic; scrambling it, as in \citet{Kolchinsky2018SemanticIA}, leaves reward at chance. We consider such variables \textit{contextual} since they mediate dependencies between action, observations, and reward without being directly causal. Here, we introduce \textbf{Factorization Regret (FR)}: the contribution of the latent variable interactions to inference of the optimal action. More generally, FR emerges when conditioning on observables changes the mutual information between contextual variables and the optimal action. Conceptually, FR measures how much more reliable a policy becomes after accounting for latent variable interactions. 

We formalize \textit{meaningful} intelligence via Information Theory and Reinforcement Learning as meta-learning from reward how to perform inference in environments with compositional latent structure. We support this view with four contributions: \textbf{(1)} A family of Partially Observable Markov Decision Processes (POMDPs) in which Factorization Regret can be manipulated (\autoref{sec:fr}). \textbf{(2)} Factorization Regret's account of the performance gap between Fully Trained Recurrent Neural Networks (RNNs) and Echo State Networks (ESNs) (\autoref{exp1}). \textbf{(3)} An architecture which explicates how latent variable interactions can be learned from naturalistic feedback (\autoref{exp2}). \textbf{(4)} An account of how learning interactions enables zero-shot control in novel settings (\autoref{exp3}).

\section{Related Works}

\subsection{Compositional Generalization} 
Animals learn to interact with the environment via the reinforcement of actions that maximize \textit{reward} given the state of the world \citep{Skinner1953ScienceAH, Bellman1957DynamicP, Sutton1998ReinforcementLA, Russell2003ArtificialI}. By learning how different elements of the world interact with each other, an animal may be able to generalize to a novel environment \citep{hayes2001relational}. Generalizing to a novel combination of elements is known as \textit{compositional generalization} \citep{Kay2023EmergentND, Lampinen2025LatentLE}, and occurs naturally in recurrent networks trained to predict complex sensory inputs \citep{boeshertz2025predictive}. For instance, the Hippocampus (HPC) has been modeled as a network that learns a compositional grid of sensory predictions \citep{Whittington2019TheTM}. Navigating this grid is considered an ``internal action'' that encodes object relationships and facilitates prediction. Similarly, models of decision-making consider sequential inference akin to navigation towards a ``decision boundary'' \citep{Shadlen2001NeuralBO, Gold2007TheNB, Kiani2009RepresentationOC, Mante2013ContextdependentCB, Brunton2013RatsAH, deLafuente2015RepresentationOA, Kang2017PiercingOC, Pereira2021EvidenceAR, Safavi2022MultistabilityPV, Safavi2025ADM}. Treating the internal `commitments' of a decision-making process as elemental units could explain how animals generalize to novel combinations of task-relevant outputs. The Cognitive Gridworld is intended as a minimal setting for studying such generalization, complementary to large-scale benchmarks such as ARC-AGI \citep{Chollet2024ARCP2}, without the need for large-scale agents such as Gato \citep{Reed2022AGA} and Dreamer \citep{Hafner2025MasteringDC}.

\subsection{Partially Observable Markov Decision Processes} Animals often learn without concrete examples or explicit symbols; for instance, riding a bike becomes automatic without a full understanding of its mechanics. Such abstract knowledge is often implicit---a `gut-feeling' that is effortlessly utilized, but hard to describe, which guides action without access to the exact state of the world. Decision-making under this kind of uncertainty is naturally modeled via POMDPs \citep{Monahan1982StateOT, Kaelbling1998PlanningAA} that utilize a \textit{belief-state}, a probability distribution over all possibilities, updated by observation or active exploration \citep{Hu2018ActiveLW, karni2025goals}. This belief-state is often intractable, requiring that relevant information be compressed into an \textit{approximate belief-state} \citep{strm1965OptimalCO, Roy2011FindingAP}. For instance, clever use of memory could store the minimal statistics sufficient for \textit{marginal} belief updating. Dopamine, which encodes belief-like reward prediction errors, could embed such statistics into the low-dimensional manifold that cortical dynamics traverse \citep{Hollerman1998DopamineNR, Mante2013ContextdependentCB, Hennig2023EmergenceOB, Churchland2024PreparatoryAA}. By using observations that lack spatiotemporal structure, we can isolate dynamics that arise purely from the intractability of the belief-state.

\subsection{Meta-learning} Animals cluster observations by independent \textit{latent causes} \citep{Gershman2010LearningLS, Dubreuil2022TheRO} via mechanisms of the brain such as attention and computation through dynamics \citep{Niv2015ReinforcementLI, Vyas2020ComputationTN, Versteeg2025ComputationthroughDynamicsBS}. Together, these form a traversable \textit{cognitive map}—an abstract space of task-relevant variables \citep{Tolman1948CognitiveMI, Behrens2018WhatIA, Garvert2023HippocampalSC}. This process can be described as meta-learning, wherein agents learn to do inference \citep{Santoro2016MetaLearningWM, Wang2018PrefrontalCA, Yang2019TaskRI, Hospedales2020MetaLearningIN, Turner2023TheSB, Lake2023HumanlikeSG, Bowler2025StructuredES, Lan2025GoaldirectedNI}. Meta-learning owes its flexibility to the compositionality of the underlying generative process. Utilizing this compositionality, \cite{Achille2018LifeLongDR} developed VASE, an algorithm that generates novel data augmentations by recombining learned generative factors. Similar approaches have been useful in Reinforcement Learning (RL) settings for few-shot, zero-shot, in-context and offline learning \citep{zintgraf2019varibad, Rakelly2019EfficientOM, chen2021decision, laskin2022context, Ajay2022IsCG}. For all its flexibility, meta-learning is formally defined in a world that is fixed but unknown \citep{Ortega2019MetalearningOS}. Yet, the complexity of meta-learning tasks hinders theoretical investigation of how dynamics \textit{learn to meta-learn}. Here, we design meta-learning tasks with a flexible generative process that remains analytically tractable.

\section{Problem Setup}

\begin{figure}[h]
  %\vspace{-2mm}
  \centering
   \includegraphics[width=1\columnwidth]{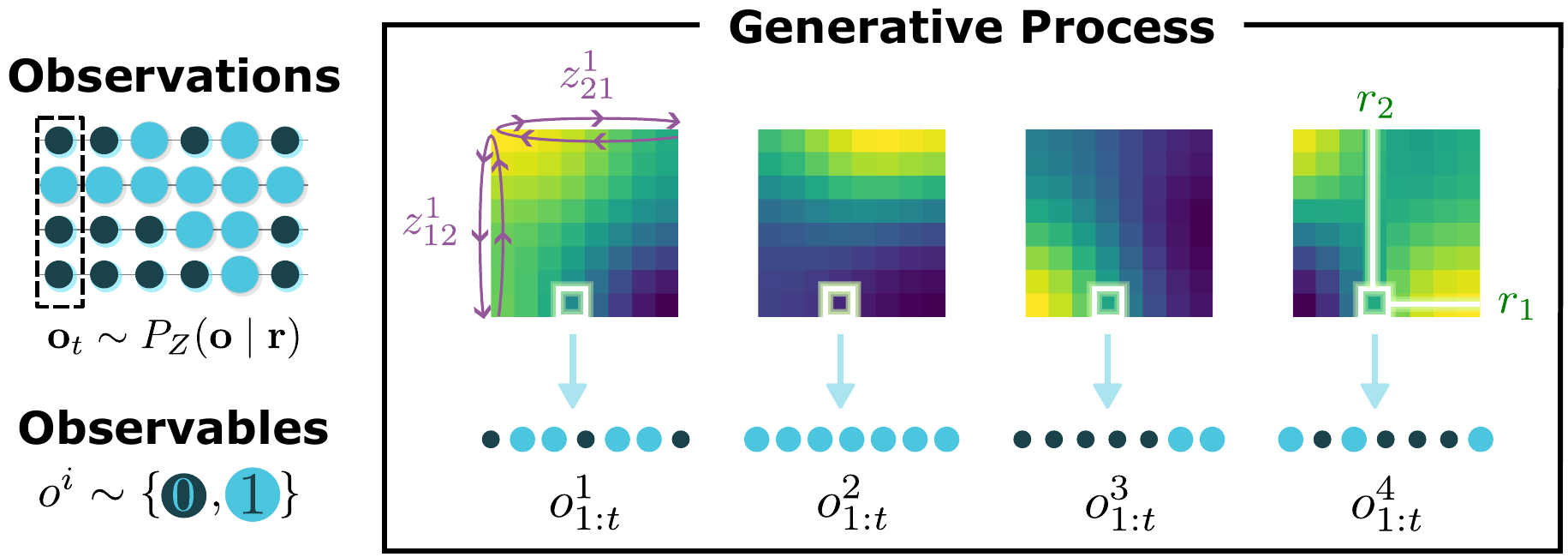}
   %\vspace{-2mm}
\caption{\textbf{Environment schematic for $C=2$.} Observations $\mathbf{o}_t$ are generated stochastically. Variable interactions $Z$ parameterize likelihood $P_Z(\mathbf{o} \mid \mathbf{r})$ and variable realizations $(r_1, r_2)$ fix the probability of sampling each observable $o^i$.}
% \caption{\textbf{Environment schematic for $C=2$.} Observations $\mathbf{o}_t$ are generated stochastically by sampling from likelihood $P_Z(\textbf{o} \mid \mathbf{r})$, parameterized by variable interactions $Z$ and realizations $\mathbf{r}$.}

  \label{fig:latent_generative}
  %\vspace{-4mm}
\end{figure}

We assume that only a few latent variables are relevant to a specific goal. We refer to the actual value of a latent variable as its \textbf{realization}, governed by a random process where each variable, in a \textbf{context} of $C$ relevant variables, takes a value from $R$ possible realizations:
\[\mathbf{r} \overset{\text{i.i.d.}}{\sim} U(\{1,\dots, R\})^{C}, \quad \text{where } r_c \in \mathbf{r} \text{ is the realization of variable } c. \]
The agent observes the world through a $d_o$-dimensional observation space $\mathcal{O}$ (\autoref{fig:latent_generative}). At time $t$, the probability of each independent observable $o^i \in \{0,1\}$ within observation $\mathbf{o}_t$ depends on realizations $\mathbf{r}$ and a likelihood parameterized by interactions $Z$. When $C=1$, variables only interact with themselves. For all $C \geq 2$ there are $|Z|=d_o \times C \times (C-1)$ pairwise interactions between latent variables. $\mathbf{z}^i \in Z$ are the interactions relevant to $o^i$. We let $\ell_{\mathbf{z}^i}(\mathbf{r})$ denote the likelihood $P_{ \mathbf{z}^i}(o^i=1 \mid \mathbf{r})$, forming a tensor of order $C$ with $R$ entries along each axis. Construction of $\ell_{\mathbf{z}^i}(\mathbf{r})$ is described in detail in \autoref{app:de}; a brief description follows.

Each pairwise interaction $z_{cc'}^i$ (the modulation from variable $c'$ onto $c$) shifts the phase of a sinusoidal base to form vectors $\mathbf{v}_{cc'}^i(r)$, functionally ``rolling'' values of $\ell_{\mathbf{z}^i}(\mathbf{r})$ along axis $c$. When $C=1$, the likelihoods are given by vector $\ell_{\mathbf{z}^i} = \sigma(\mathbf{v}^i) $, where $\sigma$ denotes element-wise sigmoid. For all $C \geq2$, broadcasting operator $\mathcal{B}_{cc'}$ entangles $r_c$ and $r_{c'}$ by mapping the $R \times R$ outer product $\mathbf{v}_{cc'}^i (\mathbf{v}_{c'c}^i)^\top$ onto the $C$-dimensional likelihood tensor $\ell_{\mathbf{z}^i}$:
\[ \ell_{\mathbf{z}^i} = \sigma\left(\sum_{c< c'} \mathcal{B}_{cc'}\left(\mathbf{v}_{cc'}^i (\mathbf{v}_{c'c}^i)^\top\right)\right)  \quad C \geq2 \]

An episode $E$ is defined by the tuple $(\mathbf{r}, g, Z, \mathbf{o}_{1:T})$, comprising the realizations, a random goal index $g \sim U(\{1,\dots ,C\})$, the interactions, and a trajectory of $T$ observations. The agent's objective is to infer the realization of goal variable $r_g$ and select optimal action $a^{\star}=r_g$ from $a \in \{1,\dots,R\}$. Optimal inference of $r_g$ can be achieved with sequential Bayesian updating:
\[P_Z(\mathbf{r}\mid \mathbf{o}_{1:t}) \propto P_Z(\mathbf{o}_t\mid \mathbf{r})\,P_Z(\mathbf{r}\mid \mathbf{o}_{1:t-1}),\]
where the joint likelihood of the independent observables follows a product of Bernoullis:
\[P_Z(\mathbf{o}_t\mid \mathbf{r})  =\prod_{i=1}^{d_o}\ell_{\mathbf{z}^i}(\mathbf{r})^{o_t^i}(1-\ell_{\mathbf{z}^i}(\mathbf{r}))^{1-o_t^i}.\]

\section{Results}

\subsection{Factorization Regret quantifies the cost of Naive Bayes}\label{sec:fr}

\begin{figure}[h]
  %\vspace{-2mm}
  \centering
   \includegraphics[width=1\columnwidth]{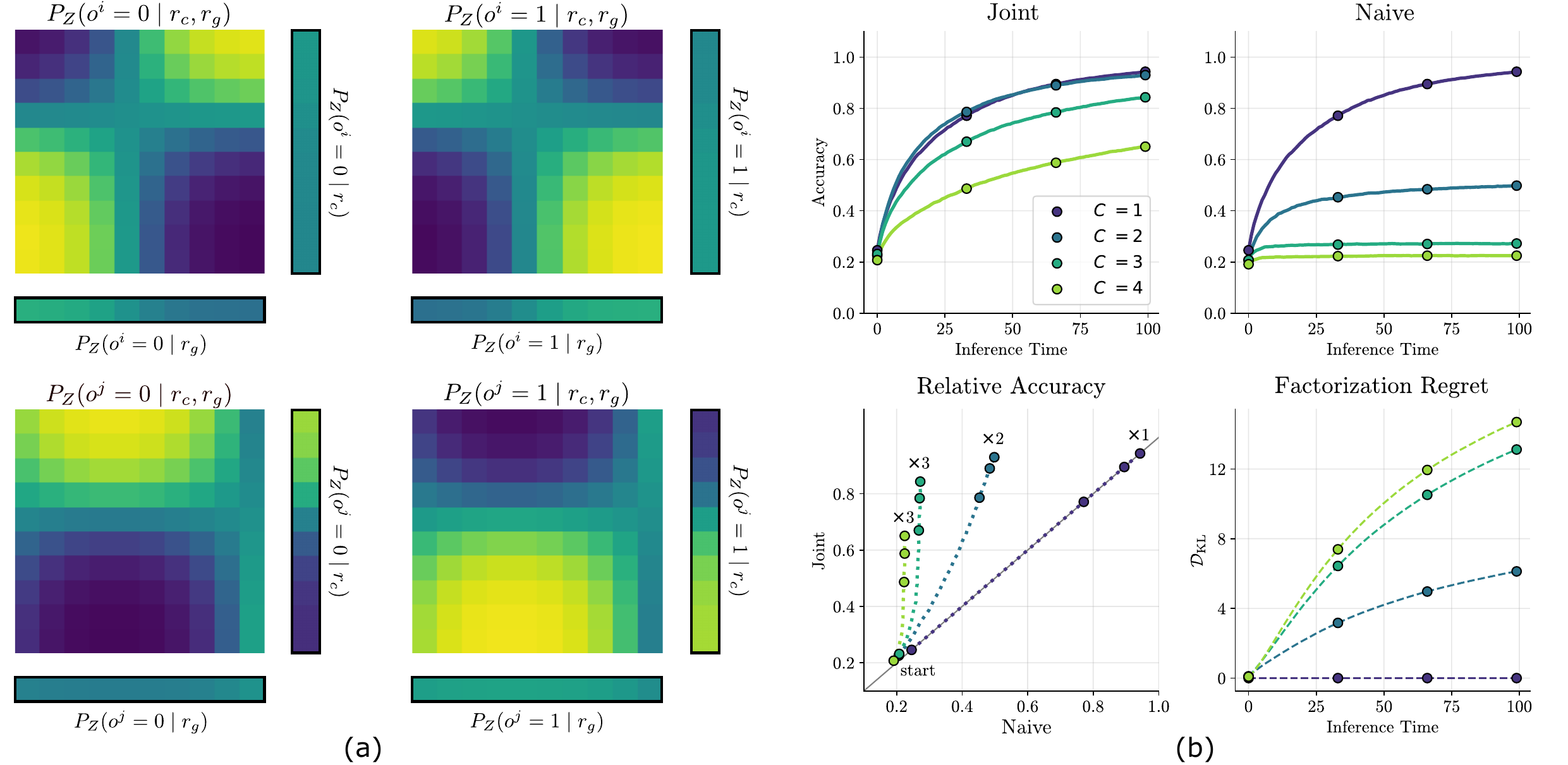}
   %\vspace{-2mm}
\caption{\textbf{The cost of Naive Bayes grows with time and interactions.} 
\textbf{(a)} Example Joint (matrices) and marginalized (vectors) likelihoods. \textbf{(b)} Top: Accuracy of Joint (left) and Naive (right) Bayes across varying context sizes. Bottom: Relative accuracy (left) and Factorization Regret (right). Circles mark four equidistant reference time-points throughout inference.}
  \label{fig:JN_comparison}
  %\vspace{-4mm}
\end{figure}

A natural baseline is a Bayesian observer that updates marginal beliefs ($B_{tc}$) independently. The likelihood for this \textit{Naive} observer is acquired by marginalizing the joint likelihood (\autoref{fig:JN_comparison}a). We quantify the contribution of latent variable interactions to task performance at time $t$ with a Kullback-Leibler divergence ($\mathcal{D}_{\mathrm{KL}}$) between the Naive and Optimal Joint Bayesian observers' marginal beliefs over the goal variable, which we refer to as Factorization Regret (derived in \autoref{app:sii_derivation}). \[\text{Factorization Regret}_t = \mathcal{D}_{\mathrm{KL}}\!  (B_{tg}^{\mathrm{Joint}}\|B_{tg}^{\mathrm{Naive}})\] FR quantifies the accumulating cost of scrambling Interaction Information on every step. 

\autoref{fig:JN_comparison}b shows that average performance (calculated as $\langle [\arg\max_r \bigl(B_{tgr}\bigr)=a^{\star}]\rangle_E$ using Iverson bracket notation) improves with time and declines with the number of interacting variables, for both observers. However, the relative accuracy of joint and naive observers grows rapidly with interactions, along with FR. In summary, we have a family of POMDPs where the gap between Joint and Naive Bayes can be manipulated and attributed entirely to latent variable interactions.

\subsection{Experiment 1: The Cognitive Gridworld requires flexible neural dynamics} \label{exp1}

To demonstrate the practical consequences of Factorization Regret, we compare Fully Trained RNNs and ESNs, commonly used as models of neural dynamics \citep{Mante2013ContextdependentCB, Wang2018PrefrontalCA, Yang2019TaskRI}. Echo State dynamics provide random temporal features for a readout to learn an arbitrary function of past inputs. Yet, they suffer from tradeoffs between discrimination versus generalization, memory versus nonlinearity, and recall versus precision \citep{alvarez2002exact, Dambre2012InformationPC, Barak2013TheSO, Inubushi2017ReservoirCB, Tio2019DynamicalSA, Tino2024PredictiveMI}. We find that Naive Bayes predicts the precise sub-optimal performance achieved by ESNs.

\paragraph{Interaction differentiates network capabilities at inference time.} 

\begin{figure}[h]
  %\vspace{-2mm}
  \centering
   \includegraphics[width=1\columnwidth]{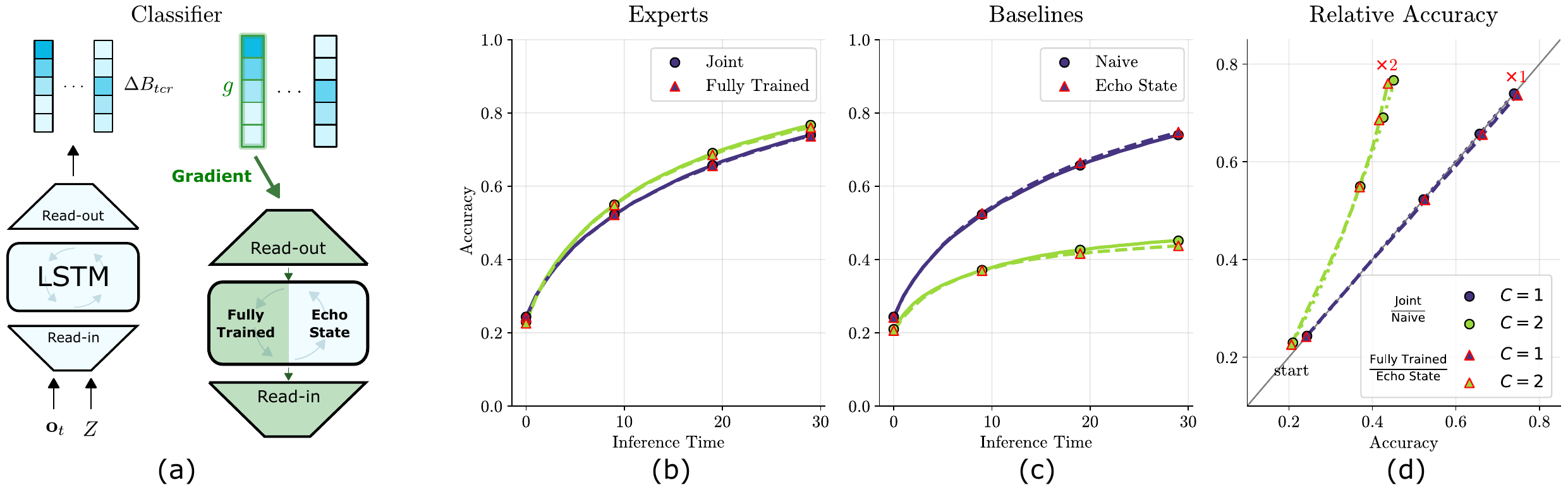}
   %\vspace{-2mm}
  \caption{\textbf{RNNs align with theoretical predictions}. \textbf{(a)} Architecture (left) and gradient flow (right) of the Classifier. Only the goal belief receives a gradient. \textbf{(b-d)} Same as \autoref{fig:JN_comparison}b for $C = 1, 2$ ($C=3$ in Supplementary \autoref{fig:more_ctx}) with Fully Trained RNNs and ESNs (averaged over $n=5$ random seeds). Markers indicate four equidistant reference time-points throughout inference.}
  \label{fig:NB_comparison}
  %\vspace{-4mm}
\end{figure}

 The network performing sequential inference, referred to as the Classifier, consists of Long Short-Term Memory (LSTM) units and receives $Z$ and $\mathbf{o}_{1:T}$. We train the network's goal belief $B_{tg}^{\text{Net}}$ to converge to $B_{tg}^{\mathrm{Joint}}$ with loss:
$$ \mathcal{L}_{\text{FR}} =  \left( \dfrac{1}{2}\mathcal{D}_{\mathrm{KL}}(B_{tg}^{\mathrm{Joint}} \,\|\, B_{tg}^{\mathrm{Net}}) + \dfrac{1}{2}\mathcal{D}_{\mathrm{KL}}(B_{tg}^{\mathrm{Net}} \,\|\, B_{tg}^{\mathrm{Joint}}) \right)^{\alpha}, $$
where $\alpha=0.5$ improves training efficiency. We use network output as marginal belief \textit{updates} $\Delta B_{t} \in \mathbb{R}^{C\times R}$ (\autoref{fig:NB_comparison}a), where $r_g$ can be estimated with a sum of log likelihood updates:
\[
\arg\max\limits_{r_g} P_Z(r_g \mid \mathbf{o}_{1:T}) = \arg\max\limits_{r_g}  \left( \sum_{\mathbf{r}_{\setminus g}} \exp \left( \sum_{t=1}^{T} \ln P_Z(\mathbf{o}_{t} \mid \mathbf{r}) \right) \right).
\]
Notice that the summation over non-goal realizations $\mathbf{r}_{\setminus g}$ envelops the temporal accumulation of a log probability. Since this cannot be decomposed into a summation over marginal likelihoods, this is Markovian only with respect to the \textit{joint} posterior. By circumventing the challenge of integration, this setup isolates the capacity of memory dynamics to \textit{re-evaluate} marginal beliefs with history-dependent corrections, which can be functionally equivalent to updating the joint posterior (\ref{meth:exp1}). In \autoref{fig:NB_comparison}b-d, the Fully Trained RNN reaches optimality, whereas the ESN, trained towards the same target, plateaus precisely at Naive Bayes. As predicted, when $C = 1$ and FR is zero, the gap between networks vanishes. The performance benefit of task-aligned dynamics \citep{Schuessler2023AlignedAO} is apparent in the relative accuracy between networks, which is identical to the ratio between Bayesian observers. These results still persist when the number of trainable parameters is matched and even when we use ESN output to update joint beliefs (Supplementary \autoref{fig:controls}).

\begin{figure}[h]
  %\vspace{-2mm}
  \centering
   \includegraphics[width=1\columnwidth]{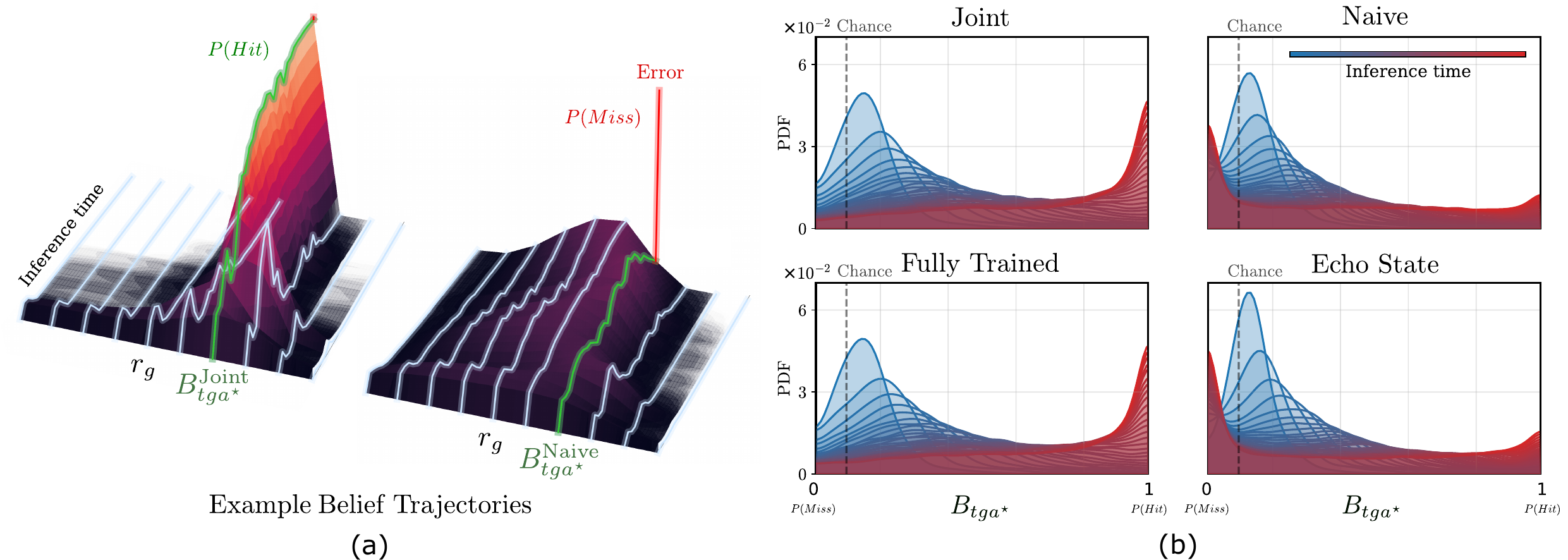}
   %\vspace{-2mm}
  \caption{\textbf{Failure to capture FR can induce hallucinations}. \textbf{(a)} Sequential updating of marginal beliefs under Joint and Naive inference. \textbf{(b)} Distribution of belief assigned to the correct action at each step of inference, pooled over episodes. Colors progress from blue to red throughout inference.}
  \label{fig:belief}
  %\vspace{-4mm}
\end{figure}

\paragraph{Naive Bayes predicts the failure mode exhibited by Reservoir Computing.} Next, we take a closer look at the belief dynamics in $C=2$, where Naive Bayes predicts a unique failure mode analogous to hallucination. \autoref{fig:belief}a shows a representative trajectory of goal belief under Joint and Naive Bayes where $P(Hit)=1-P(Miss)$ is the belief assigned to the correct action. In \autoref{fig:belief}b, we pool $P(Hit)$ over many episodes and analyze its density at each step $t$ of inference. Early on, the distribution of \textit{Hits} is centered near chance, but as time progresses (blue $\to$ red) the density shifts rightward, reflecting a growing $P(Hit)$. Interestingly, in the right column a second, disjoint peak below chance emerges. This reflects a significant portion of trajectories where misinterpretation makes evidence \textit{harmful} to Naive Bayes and ESNs, an effect that grows with FR (Supplementary \autoref{fig:calibration}). In \ref{app:B3} we explore the hypothesis that Fully Trained RNNs stay reliable by adapting their dynamics to the history-dependence of the optimal marginal belief update.

\subsection{Experiment 2: Compositional world structure can be learned indirectly}\label{exp2} 

\begin{figure}[h]
  %\vspace{-2mm}
  \centering
  \includegraphics[width=1\columnwidth]{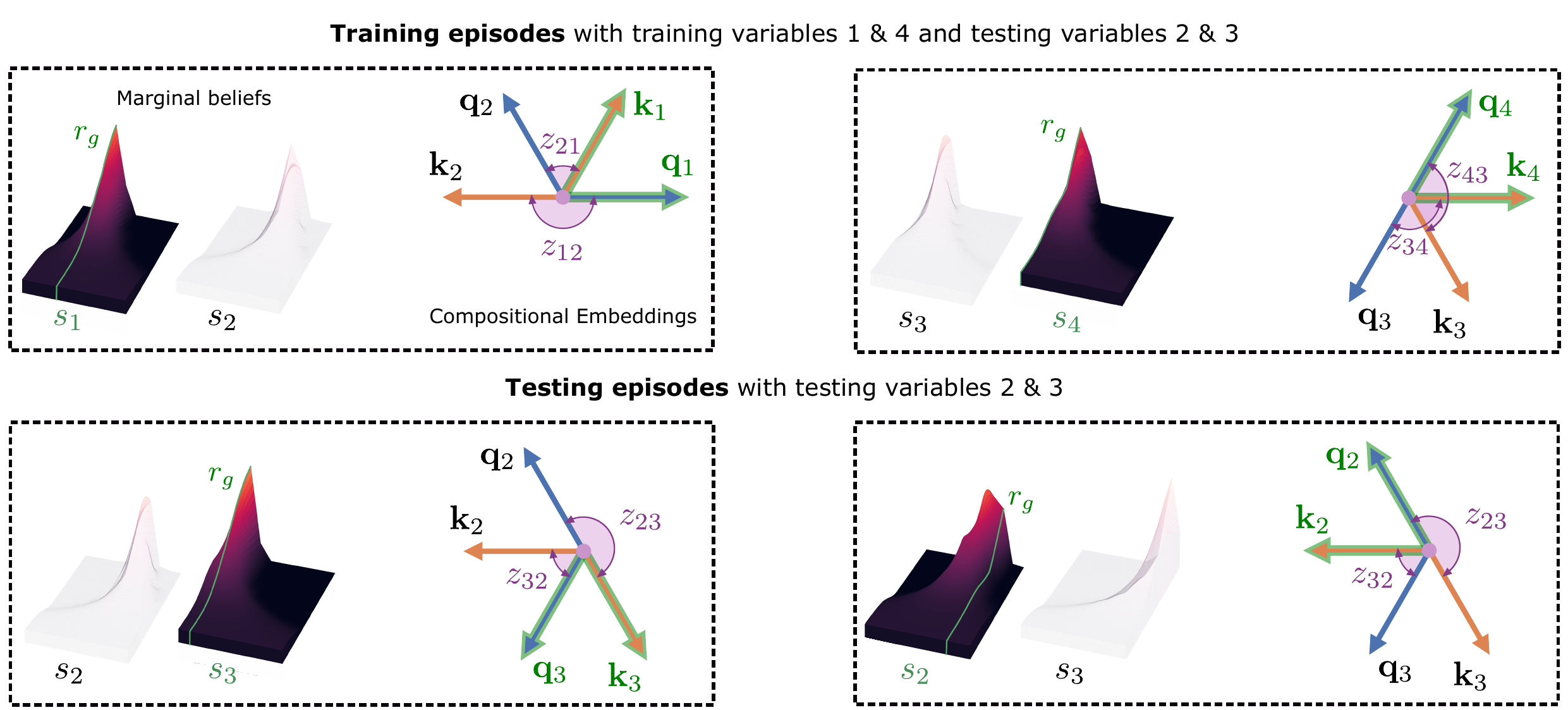}
  %\vspace{-2mm}
  \caption{\textbf{Compositional generalization schematic.} Testing variable embeddings must be learned through their implicit relationships to training variables. Goal embeddings are outlined in green.}
  \label{fig:generalization}
  %\vspace{-4mm}
\end{figure}

Thus far, we have established that neglecting Factorization Regret impedes optimal decision-making. However, networks in Experiment 1 were trained with explicit access to $Z$. Now we treat $Z$ as unknown but learnable. Consider a large set $\mathcal{S}$ of latent variables, where the possible realizations of variable $s\in\mathcal{S}$ represent a set of mutually exclusive realizations of the world, e.g. different seasons. Depending on the season, sunlight in the morning could mean that your alarm will trigger soon or that you've already slept through it. We assume that each variable carries a fixed set of vectors that encode how it interacts with other variables. Thus, learning how season interacts with time of day helps predict how season will interact with other variables. We call these vectors \textit{Compositional Embeddings}. Formally, we associate variable $s$ with a fixed set of embedding vectors $\{(\mathbf{k}_s^i, \mathbf{q}_s^i)\}_{i=1}^{d_o}\in\mathcal{E}$, and compute interactions $z_{cc'}^i=(\mathbf{k}_c^i)^\top \mathbf{q}_{c'}^i$ between all distinct pairs $c,c'$ of relevant variables in the context $\mathcal{C}\subset \mathcal{S}$ (where $|\mathcal{C}|=C$). Since our focus is generalization to unseen combinations, we treat variable identities $c,c'\in \mathcal{C}$ as known and treat $10\%$ of all variables as \textit{testing variables}. Episodes are split into \textit{training episodes}, which contain at most one testing variable (never the goal), and \textit{testing episodes}, which consist entirely of \textit{testing variables}. Since belief over a testing variable never receives a gradient, its embeddings must be learned indirectly (\autoref{fig:generalization}). 

\begin{figure}[h]
  %\vspace{-2mm}
  \centering
  \includegraphics[width=1\columnwidth]{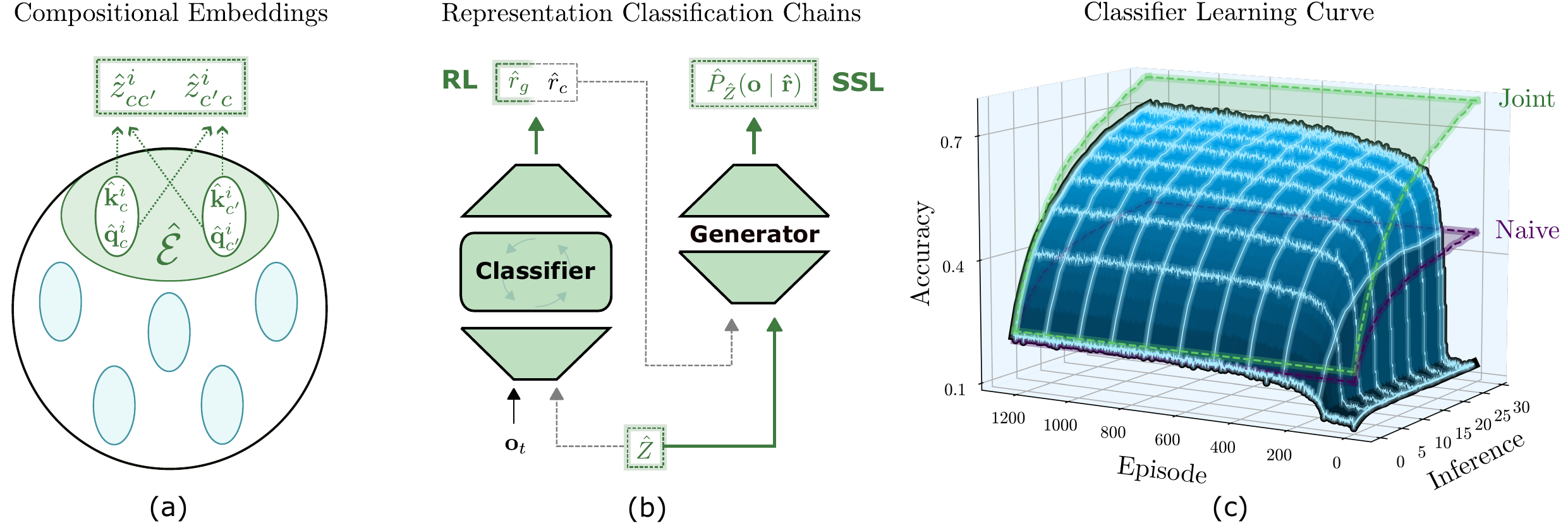}
  %\vspace{-2mm}
  \caption{\textbf{A variational architecture learns Compositional Embeddings from reward.} \textbf{(a)} Relevant variables interact via learnable embeddings to form interactions. \textbf{(b)} Forward pass and gradient flow of the Classifier and Generator. The Classifier learns from rewards while the Generator uses self-supervised learning (SSL). \textbf{(c)} Testing episode accuracy of the Classifier throughout training.}
  \label{fig:embed}
  %\vspace{-4mm}
\end{figure}

\paragraph{Naturalistic feedback is sufficient for learning interactions.} By conditioning a generative model $\hat{P}_{\hat{Z}}(\mathbf{o} \mid \hat{\mathbf{r}})$ on estimated interactions $\hat{Z}$, a gradient can be propagated back to learnable embeddings $\{(\hat{\mathbf{k}}_s^i, \hat{\mathbf{q}}_s^i)\}_{i=1}^{d_o}\in \hat{\mathcal{E}}$ (\autoref{fig:embed}a). Estimated realizations $\hat{\mathbf{r}}$ are sampled from the Classifier's final belief $B_T$ (\ref{meth:exp2}). For $\hat{P}_{\hat{Z}}(\mathbf{o} \mid \hat{\mathbf{r}})$ to be learnable,  $\hat{\mathbf{r}}$ must be reliable. This presents a dilemma: learning embeddings requires accurate inference, yet accurate inference relies on the learned embeddings. Motivated by V-JEPA 2-AC \citep{Assran2025VJEPA2S}, we separate the challenges of representation and classification. \textit{Representation Classification Chain} (RCC) (\autoref{fig:embed}b) modules bootstrap off each other without sharing gradients: a Classifier uses $\hat{Z}$ to infer $\hat{\mathbf{r}}$, while a Generator uses $\hat{\mathbf{r}}$ to reconstruct observations (\ref{meth:exp2}). The Classifier uses its goal belief as a policy, sampling one action at the end of each trajectory $a \sim B_{Tgr}$ to acquire sparse, partial, and binary feedback depending on whether or not the selected action was optimal $\delta(a-a^{\star}) \in \{0,1\}$:
$$ \mathcal{L}_{\text{Classifier}} \approx -\frac{1}{T}\sum_t \big( \delta \ln B_{tga} + (1-\delta) \ln (1-B_{tga}) \big). $$
Concurrently, the Generator minimizes a Jeffreys Divergence, $J(p \,\|\, q) =\mathcal{D}_{\mathrm{KL}}(p \;\|\;q)+\mathcal{D}_{\mathrm{KL}}(q \;\|\;p)$,
between the empirical trajectory distribution $\tilde{P}(\mathbf{o})$ and its own predictions $\hat{P}_{\hat{Z}}(\mathbf{o}\mid \hat{\mathbf{r}})$:
$$ \mathcal{L}_{\text{Generator}} = J\bigl(\tilde{P}(\mathbf{o}) \,\|\, \hat{P}_{\hat{Z}}(\mathbf{o}\mid \hat{\mathbf{r}})\bigr), $$
where the gradient flows from $\hat{Z}$ to $\hat{\mathcal{E}}$, teaching the embeddings to be useful for prediction. \autoref{fig:embed}c demonstrates that $\hat{\mathcal{E}}$ is indeed learnable from naturalistic feedback. Interestingly, this generalization only emerges once $|\mathcal{S}|$ is large enough. When the world contains too few latent variables, training variable embeddings are learned, but this does not transfer to testing variables (Supplementary \autoref{fig:emb}). Ablating the Generator and using gradient flow from the Classifier significantly impairs learning (Supplementary \autoref{fig:ablation}). 

\subsection{Experiment 3: Latent space navigation enables zero-shot control}
\label{exp3}

\begin{figure}[h]
  %\vspace{-2mm}
  \centering
 \includegraphics[width=1\columnwidth]{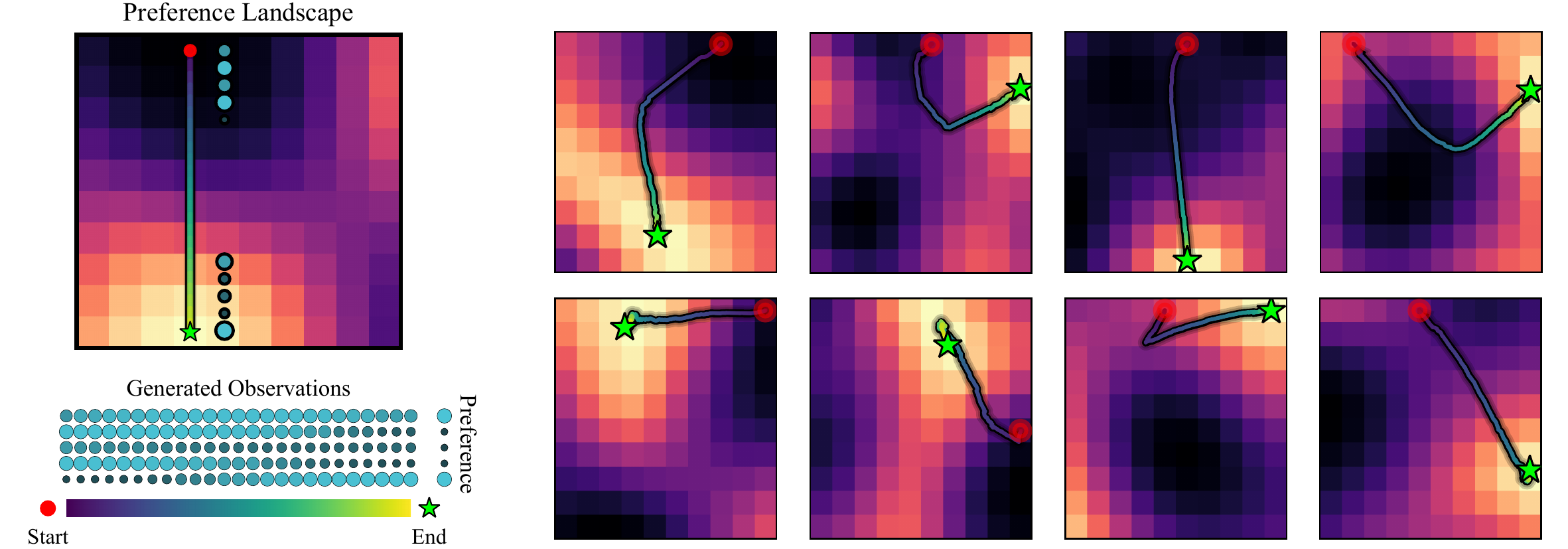}
 %\vspace{-2mm}
  \caption{\textbf{Latent space navigation enables zero-shot control over observations.} Schematic illustration (left) and empirical trajectories (right) of an agent navigating a novel latent space to identify the state of the world which would generate the maximally preferred observation. Heatmaps indicate $\mathcal{R}(\mathbf{r}, \mathbf{\Omega}, \ell)$ and trajectories show $\arg\max_{\mathbf{r}}\pi(\mathbf{r} \mid \hat{Z})$ throughout training (smoothed for clarity).}
  \label{fig:RL2}
  %\vspace{-4mm}
\end{figure}

There is a developing interest in the generalization that emerges from conditioning on variables such as action, context, cluster, class, and constraints \citep{Hu2018ActiveLW, Rakelly2019EfficientOM, Ajay2022IsCG, Hafner2025MasteringDC}. In the previous section, we demonstrated that conditioning on interactions enables generalization to novel combinations of latent variables. Building on this, we show how learned Compositional Embeddings facilitate optimal behavior at the first opportunity for control, such as using a brand new oven. In this setup we consider an agent that has control over the latent variables $\mathbf{r}$, modeling a hypothetical interaction with the world. Note that the action space has expanded to the entire latent space. A \textit{preferred observation} $\mathbf{\Omega}$ indicates success (such as the smell of a freshly baked pizza) and associates each location in latent space with \textit{intrinsic reward} $\mathcal{R}$:
\[\mathcal{R}(\mathbf{r}, \mathbf{\Omega}, \ell) = \exp\left( \frac{1}{d_o}\sum_i \bigl(\ln(\ell_{\mathbf{z}^i}(\mathbf{r})) \Omega^i + \ln(1-\ell_{\mathbf{z}^i}(\mathbf{r}))(1-\Omega^i)\bigr) \right).\]
 We can visualize the dependence of $\mathcal{R}(\mathbf{r}, \mathbf{\Omega}, \ell) $ on $\mathbf{r}$ as a \textit{preference landscape} where the task is to find its maxima--- the action most likely to generate a preferred observation (\autoref{fig:RL2}, left). For a given $\mathbf{\Omega}$, a simple (and memory-less) controller can optimize a context-dependent policy $\pi(\mathbf{r} \mid \hat{Z})$ and value function $\nu(\hat{Z})$, with a standard Advantage Actor-Critic (A2C) loss:
$$ \mathcal{L}_{A2C} \approx A^2 - A_{\text{Detached}}\ln\pi(\mathbf{r} \mid \hat{Z}) +\beta_{cntrl}\textstyle\sum_{\mathbf{r}'}\pi(\mathbf{r}' \mid \hat{Z}) \ln \pi(\mathbf{r}' \mid \hat{Z}), $$
where advantage $A = \mathcal{R}(\mathbf{r}, \mathbf{\Omega}, \ell) - \nu(\hat{Z})$ and $\beta_{cntrl}$ is an entropy bonus (see \autoref{meth:exp3}).

\paragraph{Training in imagination.} Notice that intrinsic reward $\mathcal{R}(\mathbf{r}, \mathbf{\Omega}, \ell)$ depends on $\ell$. The RCC can approximate $\ell$ with its learned $\hat{P}_{\hat{Z}}(\mathbf{o}\mid \mathbf{r})$, which allows the controller to be trained offline in imagined environments. This inverts offline model-based optimization \citep{Trabucco2021ConservativeOM}, which learns the objective function from a fixed set of evaluated observations. In our setting, the RCC imagines and evaluates observations to help a controller navigate latent space. \autoref{fig:RL2} (right) displays several example trajectories of $\arg\max_{\mathbf{r}}\pi(\mathbf{r} \mid \hat{Z})$ throughout learning, overlaid on the true preference landscape, showing that the controller identifies the optimal action without any interactions with the environment. Despite offline optimization's high sensitivity to model errors \citep{Ha2018RecurrentWM, Trabucco2021ConservativeOM}, offline training with $\ell$ and $\hat{P}_{\hat{Z}}(\mathbf{o}\mid \mathbf{r})$ is indistinguishable (\ref{app:offline_details}). 

\section{Discussion}

Interactions push us to discover more than just our goals, yet their combinatorial growth renders the world intractable. We introduced Factorization Regret to measure the goal-relevant (\textit{semantic}) information that is lost when interactions are ignored, and the Cognitive Gridworld as a setting that generates FR while remaining tractable for analytical treatment. FR explained the performance gap between Fully Trained RNNs and ESNs, and predicted a failure mode that decoupled confidence from accuracy. Because Naive Bayes is the best one can do before learning how variables interact, it offers a principled threshold for \textit{meaningful} intelligence: crossing this threshold means an agent has used its internal representation to move beyond a stimulus-bound strategy (\autoref{app:B2}).

Since FR is a property of the environment, it could quantify how much of an existing benchmark's difficulty lies in learning interactions. In other compositional benchmarks, such as Alchemy \citep{wang2021alchemy} and CompoSuite \citep{mendez2022composuite}, task structure is explicit or discovered by manipulating the world, which leaves open how much their difficulty stems from challenges beyond compositionality. In the Cognitive Gridworld all learnable structure is latent, and it can be learned without causal manipulations. This allows us to quantify the task difficulty that stems from our intended source. We presented RCCs as an existence proof that the task's compositional latent structure can be learned from naturalistic feedback, and despite our emphasis on latent space, we also demonstrate that this setting is relevant to zero-shot control.

However, the scope of both the Cognitive Gridworld and RCCs is currently limited. The spatiotemporal richness of sensory observations was abstracted away entirely, and could be reintroduced through simple latent variable dynamics. Interactions between variables could be made richer by introducing sparsity to compositional embeddings, and our tensor likelihood construction extends naturally to multiple frequencies in a discrete Fourier series. RCCs were only evaluated in small ($C=2$) environments and their scalability remains unknown. Future work should also implement additional sequence models, such as Transformers, which could vary in their mechanistic implementation of history-dependent belief updates (see \autoref{app:B3} for a preliminary investigation into the RNN implementation). The most pressing limitation is the assumption that all task-relevant variables are known, which a latent variable discovery algorithm (e.g. VASE) could solve.

Beyond artificial intelligence, this framework may also help explain how animals acquire reliable intuitions about the world. Assuming known variables and a stationary world was motivated by the hypotheses that the HPC discovers latent and temporal structure \citep{Whittington2019TheTM, Raju2024SpaceIA}. In this work, we focused on a complementary nuance: inference over interacting variables in a fixed, but unknown, environment. We propose that handling latent variable interactions is a major function of the Orbitofrontal Cortex (OFC), a region that is highly specialized in primates \citep{PreussEvolutionOP}. The OFC is involved in latent state inference, context-dependent behavior and compositionality \citep{Duarte2010OrbitofrontalCI, Takahashi2013NeuralEO, Wilson2014OrbitofrontalCA, Farovik2015OrbitofrontalCE, Schuck2016HumanOC, Chan2016APD, Pelletier2019ACR, Stalnaker2021OrbitofrontalSR, Mizrak2021TheHA, Schiereck2024NeuralDI}. Its activity is thought to encode a ``common currency'' value, including the value of information and imagined outcomes, the location of spatial goals, and a multidimensional abstract space with reward-dependent compression \citep{PadoaSchioppa2008TheRO, PadoaSchioppa2009RangeAdaptingRO, Kahnt2010TheNC, Takahashi2013NeuralEO, Stalnaker2015WhatTO, Basu2021TheOC, MuhleKarbe2023GoalseekingCN, Qiu2024FormingCM, Bussell2024RepresentationsOT}. Damage to the region has long been associated with impaired cognitive flexibility and re-evaluation \citep{Butter1969PerseverationIE, Dias1996DissociationIP, Schoenbaum2002OrbitofrontalLI, McAlonan2003OrbitalPC, Chudasama2003DissociableCO, Hornak2004RewardrelatedRL, Izquierdo2004BilateralOP, Rygua2010DifferentialCO, Rudebeck2013PrefrontalMO, Murray2015SpecializedAF}, particularly when value must be inferred, but not when cached values suffice \citep{Jones2012OrbitofrontalCS}. These seemingly disparate roles in inference and evaluation would be unified by the objective of managing Factorization Regret.

The HPC and OFC are thought to maintain complementary and interacting cognitive maps \citep{Wikenheiser2016OverTR, Wikenheiser2017SuppressionOV, Knudsen2020ClosedLoopTS, Wang2020InteractionsBH, Mizrak2021TheHA, Lin2024HippocampalAO}. The HPC is well suited to dynamic environments in which the order of events matters \citep{Raju2024SpaceIA}. We suggest that the HPC compresses ordered sensory sequences into Compositional Embeddings that are consolidated throughout the cortex. Yet once the spatiotemporal richness of the sensory world has been abstracted away, the task of belief updating still remains. Given a novel goal, the brain could recruit distributed cortical representations into a cognitive map sufficient for approximate inference of the optimal action, i.e., meta-learning. We propose that the OFC orchestrates this process, using its robust connections with limbic, thalamic, and neuromodulatory systems to influence inter-cortical communication \citep{RempelClower2007RoleOO, gao2007functional, Takahashi2009TheOC, Lodge2011TheMP, Jo2016PrefrontalRO, Namboodiri2019SinglecellAT, Weitz2019ThalamicIT} and evaluate imagined outcomes (as in \autoref{exp3}). Furthermore, by detecting unresolved Factorization Regret, the OFC could identify when new interactions should be learned by the HPC. We therefore hypothesize that detecting and managing Factorization Regret is a collaborative objective of the OFC and HPC: the HPC consolidates latent variable embeddings throughout the cortex, while the OFC recruits these representations and expands their interactions into a brain-wide cognitive map for zero-shot control.

The barrier to generalization once all relevant variables are known is understanding how to put them together. The presence of Factorization Regret indicates that learning to infer the latent state of the world could be helpful for more than just the immediate goal, and therefore environments with greater FR exert greater pressure to develop compositional representations. Rather than describing intelligence as a computation to solve a problem, this perspective casts intelligence as a property selected for by the environment. This then begs the question: what kinds of environments selected for \textit{meaningful} intelligence in animals, and can we build them?

\subsubsection*{Code Availability}
\begin{itemize}
    \item \textbf{Paper Replication:} Code to reproduce the experiments presented in this paper is available at \url{https://github.com/johnschwarcz/CognitiveGridworld}.
    \item \textbf{Environment:} A lightweight, easily customizable Cognitive Gridworld environment is available at \url{https://github.com/johnschwarcz/coggrid}.
    \item \textbf{RCC Architecture:} A lightweight, easily customizable Pytorch implementation of Representation Classification Chains is available at \url{https://github.com/johnschwarcz/rcc}.
\end{itemize}

\subsubsection*{Reproducibility Statement}
Detailed environment and model hyperparameters, extended generative process descriptions, network architectures, and pseudo-code are provided in the Appendices. The full code used to generate the results of this paper is publicly available in the repositories listed above.

\subsubsection*{Ethics Statement}
This work involved no human subjects, personal data, or deployed systems.

\bibliography{iclr2027_conference}
\bibliographystyle{iclr2027_conference}
\appendix \newpage 

\section{Methodology}

\subsection{Environment Hyperparameters}
\label{app:AEE}

\begin{itemize}
  \item $T$ (Trajectory / inference steps): 30
  \item $d_{\mathcal{E}}$ (Embedding dimensionality): 30
  \item $|\mathcal{S}|$ (Total latent variables): 500
  \item $R$ (Possible realizations): 10
  \item $d_o$ (Observation dimensions): 5
  \item $\lambda$ (Likelihood temperature): 2
\end{itemize}

\subsection{Generative process details}
\label{app:de}

Compositional Embeddings are acquired by sampling $\mathbf{k}^i_s\sim\mathcal{N}(0,1)^{d_{\mathcal{E}}}, \quad \mathbf{q}^i_s\sim\mathcal{N}(0,1)^{d_{\mathcal{E}}},\quad \forall s \in \mathcal{S}$, orthogonalizing across $\mathcal{O}$ via a Gram–Schmidt process and finally normalizing to 1 along $d_{\mathcal{E}}$. In each episode $E$, the generative process is described by:
\[o^i \sim P(o^i=1\mid\mathbf{k}^i_{1},\mathbf{q}_{{1}}^i,\dots,\mathbf{k}^i_{{C}},\mathbf{q}_{{C}}^i,r_1,\dots,r_{C}), \quad \forall i \in \mathcal{O},\]
which can be shortened to:
\[\quad \ell_{\mathbf{z}^i}(\mathbf{r}) =\mathrm{generation}(\underbrace{\mathrm{compression}(\mathbf{k}^i_{{1}},\mathbf{q}_{{1}}^i,\dots,\mathbf{k}^i_{{C}},\mathbf{q}_{{C}}^i)}_{ \underbrace{\mathrm{expansion}(z_1^{i},\dots,z^i_{C\times (C-1)})}_{\ell_{\mathbf{z}^i}}}, r_1,\dots,r_{C})\]

Generalization to an \textit{out of distribution} set $\{\mathbf{k}^i_{1},\mathbf{q}_{1}^i,\dots,\mathbf{k}^i_{{C}},\mathbf{q}_{{C}}^i\}$ may occur if the resultant $\mathbf{z}^i$ falls within the manifold of training $\mathbf{z}^i$. The low rank structure of $\mathbf{z}^i \in \mathbb{R}^{C\times(C-1)}$ is embedded in joint likelihood, $\ell_{\mathbf{z}^i} \in [0,1]^{{R}^{C}}$. Below we step through the default expansion from $\mathbf{z}^i \to \ell_{\mathbf{z}^i}$ in 2 steps.

\subsubsection*{\textbf{Step 1.} Expanding $z_{cc'}^i\to \mathbf{v}^i_{cc'}$:}
First, we expand each scalar $z^i_{cc'}$ to vector $\mathbf{v}^i_{cc'}$ through $\Theta$ and $\omega$:
\[
\mathbf{v}_{cc'}^i(r) = \sum_{n=0}^{L-1} \Theta(n,r)\ \omega(n),
\quad L = 1 + 2R.
\]
Here, \( \Theta\) is a cyclic shift of sinusoidal base values (indexing realizations from $r=0$, a relabeling of the main text's $\{1,\dots,R\}$):
\[
 \Theta(n,r) = \lambda\sin\!\biggl( \frac{2\pi}{L}\bigl(((n-r) \bmod L) + 1 - \frac{L}{2}\bigr)\biggr),
\quad \lambda \ \text{= likelihood temperature},
\]
and \(\omega\) gives softmax weights over circular distances from \(L z_{cc'}^i\):
\[
\omega(n) = \mathrm{softmax}_n\!\left( -d(n)^2 \right),
\qquad d(n) = \min\!\big(m(n),\, L - m(n)\big),
\]
\[
\text{where} \quad m(n) = \left(n - L z_{cc'}^i\right) \bmod L.
\]

Functionally, $z_{cc'}^i$ performs a smooth approximation of the discrete \textit{roll} operation of a sinusoid. Note that since $L>R$, $\mathbf{v}_{cc'}^i$ is shorter than a full period, allowing for asymmetry in $\mathbf{v}_{cc'}^i$.

\subsubsection*{\textbf{Step 2.} Expanding $\mathbf{v}_{cc'}^i(\mathbf{v}_{c'c}^i)^\top \to \ell_{\mathbf{z}^i} \in [0,1]^{R^{C}}$:}

Next, we expand outer products $\mathbf{v}_{cc'}^i(\mathbf{v}_{c'c}^i)^\top$ to tensor $\ell_{\mathbf{z}^i} \in [0,1]^{R^{C}}$ through:
\[ \ell_{\mathbf{z}^i}\;=\;
\begin{cases}
\sigma\!\big(\mathbf{v}_{cc}^i\big), & C=1, \\[6pt]
\sigma\!\left(\sum_{c< c'} \mathcal{B}_{cc'}\!\Big(\mathbf{v}_{cc'}^i (\mathbf{v}_{c'c}^i)^\top\Big)\right), & C\ge 2,
\end{cases}
\]
where $\sigma$ denotes element-wise sigmoid and $\mathcal{B}_{cc'}:\mathbb{R}^{R\times R}\to\mathbb{R}^{R^{C}}$ is a broadcasting operator that expands $\mathbb{R}^{R\times R}$ to $\mathbb{R}^{R^{C}}$ by adding singleton dimensions along all axes but $c,c'$ of $\ell_{\mathbf{z}^i}$. For example, for the case of $C=3:$ 
\[\ell_{\mathbf{z}^i} = \sigma \bigl(  \{\mathbf{v}^i_{12} (\mathbf{v}^i_{21})^\top\}_{12}   +   
\{\mathbf{v}^i_{13} (\mathbf{v}^i_{31})^\top\}_{13}    +   
\{\mathbf{v}^i_{23} (\mathbf{v}^i_{32})^\top\}_{23}  \bigr).\]
Materializing $\ell_{\mathbf{z}^i}$ for all $i$ therefore requires $O(d_o R^{C})$ memory, and each exact joint Bayesian update requires $O(d_o R^{C})$ operations. For the network experiments ($R=10$, $C\leq 3$) this is at most $5\times 10^{3}$ entries ($5\times 10^{4}$ for the exact Bayesian analysis at $C=4$ in \autoref{fig:JN_comparison}), but the exponential growth in $C$ is what renders the joint posterior intractable in general and motivates approximate, marginal belief states. By contrast, the Classifier receives $d_oC(C-1)$ interactions (for $C\geq 2$) and outputs $C\times R$ marginal logits, and the Generator's inputs grow at most quadratically in $C$, so both scale polynomially; the exponential cost is confined to the likelihood tensor, exact joint inference, and the Controller's joint policy.

\begin{figure}[H]
  \centering
   \includegraphics[width=1\columnwidth]{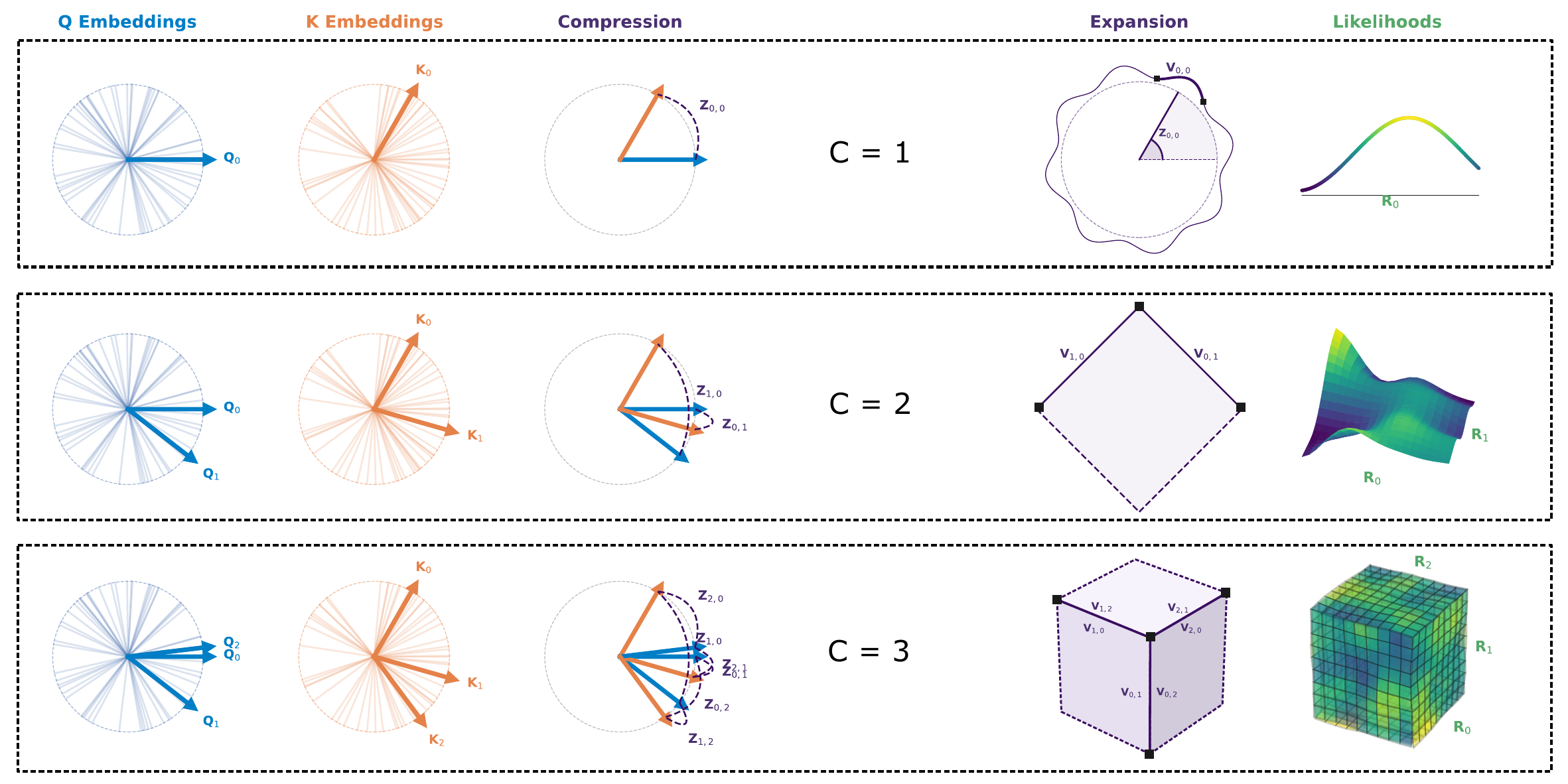}
\caption{\textbf{A flexible process for embedding Gridworld structure into latent space.} The embeddings of relevant variables are compressed into interactions which are then expanded to a discrete probability distribution over possible realizations of the world. The full process consists of first \textit{(i) compressing} embedding vectors to their scalar interactions. Then \textit{(ii) expanding} pairwise interactions to pairs of vectors, \textit{expanding} the vector pairs to matrices through an outer product, and (for $C\geq3$) \textit{expanding} the matrices into a tensor by broadcasting each matrix over all other dimensions. Finally, a valid \textit{(iii) likelihood} is ensured by a sigmoid transformation for normalization.}
  \label{fig:interact}
\end{figure} 

\subsection{Experiment 1:}
\label{meth:exp1}

\paragraph{Network goal belief.}
The goal belief $B^{\text{Net}}_{tgr}$ is extracted from the accumulated output $\mathcal{M}_{tcr} = \sum_{t'=1}^tM_{t'cr}$ (where $M_{t'cr}$ are the single-timestep network pre-activations) via a softmax over possible realizations $B^{\text{Net}}_{tgr} =\frac{e^{\mathcal{M}_{tgr}}}{\sum_{r'} e^{\mathcal{M}_{tgr'}}}.$  Although observations are conditionally independent given the realizations $\mathbf{r}$, the optimal estimate of $r_g$ requires marginalizing the joint likelihood over the non-goal realizations $\mathbf{r}_{\setminus g}$ before maximizing:
\[
\arg\max\limits_{r_g} P_Z(r_g \mid \mathbf{o}_{1:T}) = \arg\max\limits_{r_g} \left( \sum_{\mathbf{r}_{\setminus g}} \exp \left( \sum_{t=1}^{T} \ln P_Z(\mathbf{o}_{t} \mid \mathbf{r}) \right) \right).
\]
Crucially, because the summation over $\mathbf{r}_{\setminus g}$ envelops the temporal accumulation, the log-likelihood cannot be decomposed into a simple iterative summation over time for the marginals. Therefore, sequential Bayesian updating is only Markovian with respect to an intractable \textit{joint} posterior $P_Z(\mathbf{r} \mid \mathbf{o}_{1:t})$. Since $\Delta B_{tc}$ is restricted to marginal beliefs to remain tractable, re-evaluating the marginal $B_{tc}$ in light of a new observation $\mathbf{o}_{t+1}$ will not be Markovian. Therefore, performing this update accurately requires the network's memory state to encode the missing joint dependencies—specifically, the interaction information—that is lost when only tracking marginals.

\paragraph{Classifier loss.} To provide a rich teaching signal, we minimize a symmetrized KL divergence between the network and optimal goal belief-state (where $\alpha=0.5$ gave the best results):
\[
\mathcal{L}_{\text{FR}} = \left(\frac{1}{2}\mathcal{D}_{\mathrm{KL}}(B_{tg}^{\mathrm{Joint}} \,\|\, B_{tg}^{\mathrm{Net}}) + \frac{1}{2}\mathcal{D}_{\mathrm{KL}}(B_{tg}^{\mathrm{Net}} \,\|\, B_{tg}^{\mathrm{Joint}})\right)^{\alpha}.
\]

\begin{figure}[h!]
  \centering
 \includegraphics[width=1\columnwidth]{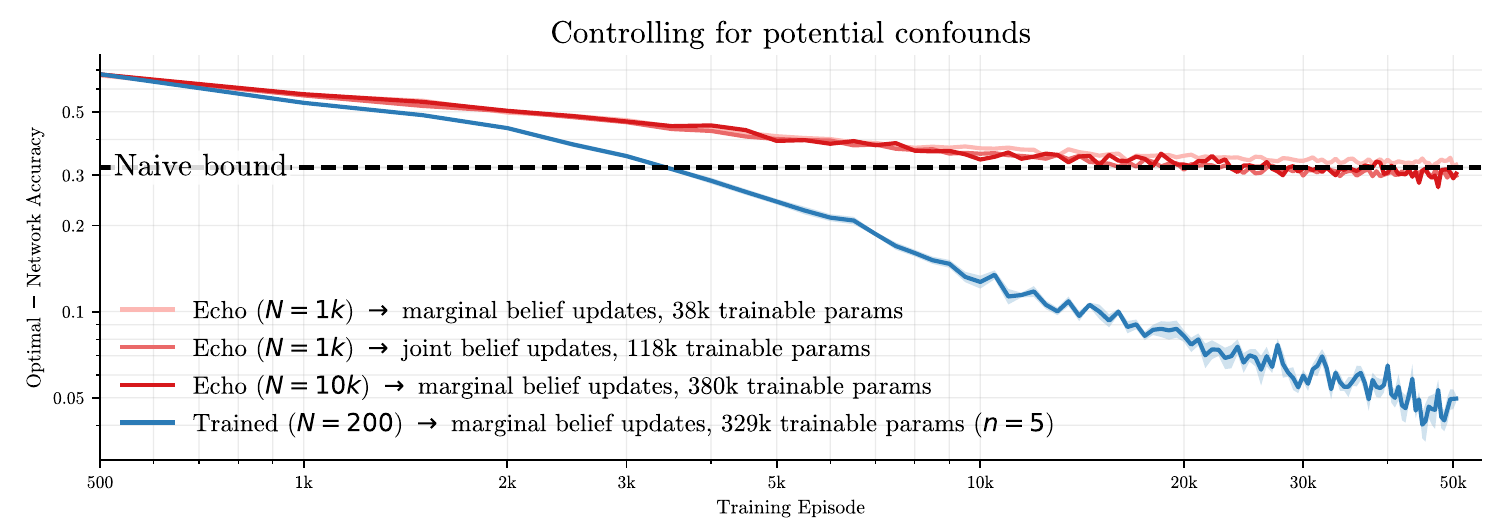}
\caption{\textbf{Results survive additional controls ($C=2$).} Excess error above optimality at the final step of inference, plotted against training episode on log–log axes. The mean ($\pm $ SEM) training curve of $5$ small Fully Trained RNNs is shown in blue. ESNs are colored light to solid red by their number of trainable parameters, where $N=10{,}000$ matches the Fully Trained RNN with $N=200$ and another ESN ($N=1{,}000$) outputs updates to the joint belief-state.}
    \label{fig:controls}
\end{figure}

\begin{figure}[h!]
  \centering
 \includegraphics[width=1\columnwidth]{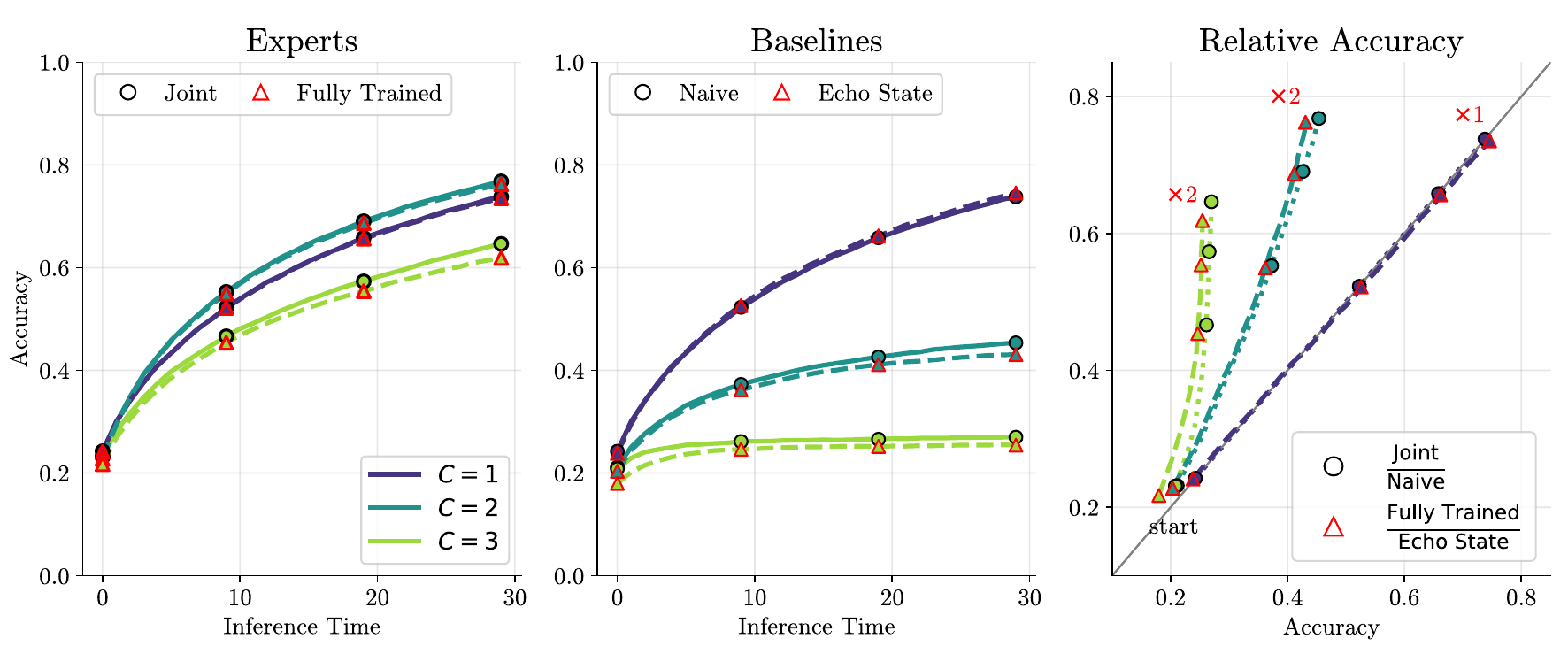}
\caption{\textbf{Results extend to 3 interacting variables.} Same as \autoref{fig:NB_comparison} with $C =3$ ($n=1$).}
    \label{fig:more_ctx}
\end{figure}

\begin{figure}[h!]
  \centering
 \includegraphics[width=1\columnwidth]{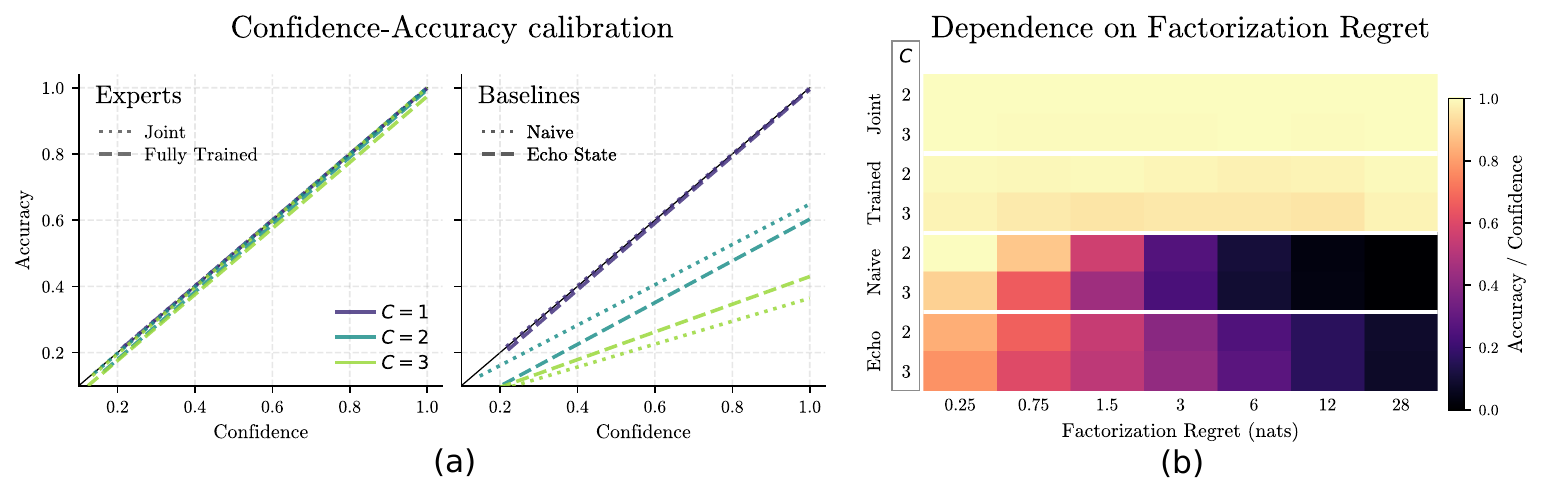}
  \caption{\textbf{Confidence--accuracy calibration for $C = 1,2,3$.} \textbf{(a)} Linear fits between confidence in the selected action and accuracy. \textbf{(b)} Accuracy over confidence as a function of Factorization Regret.}
  \label{fig:calibration}
\end{figure}

\paragraph{Confidence--accuracy calibration.}
We define an agent's confidence as the belief it assigns to its selected action, $B_{Tga}$ with $a=\arg\max_r B_{Tgr}$, and its accuracy as the fraction of episodes in which that action was correct. \autoref{fig:calibration}a fits accuracy linearly against confidence for $C=1,2,3$: Joint Bayes and the Fully Trained RNN lie on the identity line, whereas Naive Bayes and the ESN are increasingly over-confident as $C$ grows. \autoref{fig:calibration}b splits episodes by Factorization Regret and plots the ratio of accuracy to confidence. The degree of misalignment grows with the number of interacting variables in agents that neglect Factorization Regret.

\subsection{Experiment 2:}
\label{meth:exp2}

\paragraph{Classifier loss.}
At the end of each trajectory we sample an action $a\sim B_{Tgr}$ and receive outcome
$\delta(a-r_g)\in\{0,1\}$.
We train using a regularized cross-entropy-style objective:
\[
\mathcal{L}_{\text{Classifier}}
=
\dfrac{-1}{T}\sum_t
\Bigl(
\underbrace{\delta(a-r_g)}_{\text{rewarded}} \ln B_{tga}
+
\underbrace{\bigl(1-\delta(a-r_g)\bigr)}_{\text{unrewarded}}\ln(1-B_{tga})
-
\beta_{1}\underbrace{B_{tga}\ln B_{tga}}_{\text{regularization}}
\Bigr),
\]
Gradients from $\mathcal{L}_{\text{Classifier}}$ update only the Classifier parameters. Actions are sampled ($a\sim B_{Tgr}$) during training; at evaluation, the action is taken deterministically as $a=\arg\max_{r} B_{Tgr}$.

To provide the Generator estimates of the relevant variable realizations we sample $\mathbf{\hat{r}}$ from the Classifier's final belief-state $B_{T}$. We assume that the Classifier has already acted and insert the taken action $a$ into the $g$'th index of $\hat{\mathbf{r}}$.  
We also provide the Classifier's confidence in the sampled realizations $\hat{\mathbf{r}}_{conf}=B_{T\hat{\mathbf{r}}}$. The $g$'th index of $\hat{\mathbf{r}}_{conf}$ is set to $1$ if $a$ was rewarded, $0$ otherwise. 

\paragraph{Generator loss.}
We sample a latent configuration $\hat{\mathbf{r}}$ from the Classifier belief at $t=T$ (\emph{without} a gradient),
and train the Generator to maximize the likelihood of observations.
Since the true generative distribution is unknown, we construct an empirical target distribution $\tilde{P}(\mathbf{o})$, where the likelihood of each independent observable is given by its trajectory average $\tilde{P}_Z(o^i=1) = \langle o^i \rangle_t$. We use this to compute a self-supervised (since the target is constructed from observations) Jeffreys Divergence loss: 
\[ \mathcal{L}_{\text{Generator}}
=
J\Bigl(\tilde{P}(\mathbf{o})  \,\|\, \hat{P}_{\hat{Z}}(\mathbf{o}\mid \hat{\mathbf{r}})\Bigr).\]

The gradient from $\mathcal{L}_{\text{Generator}}$ updates the agent's Compositional Embeddings $\{ \hat{\mathbf{k}}_c^i, \hat{\mathbf{q}}_c^i \}_{i=1}^{d_o} \in \hat{\mathcal{E}}, \quad \forall c \in \mathcal{C}$ via gradient flow through $\hat{Z}$.

Training reliability was improved by augmenting the Generator's loss function to give greater weight to elements of a batch where the Classifier's action was rewarded. The augmented loss was: 
\[\mathcal{L}^{\star}_{\text{Generator}} = \frac{\text{chance}}{d_o}\times \mathcal{L}_{\text{Generator}} +\frac{1-\text{chance}}{\langle \delta(a-r_g)\rangle}\times \mathcal{L}_{\text{Generator}}^{\delta} + \mathcal{L}_{\text{reg}}, \]
where $\text{chance}=\frac{1}{R}$, $\mathcal{L}_{\text{Generator}}$ is averaged over all elements of the batch, $\mathcal{L}_{\text{Generator}}^{\delta}$ is the same average with unrewarded elements set to zero (so that the second term averages over rewarded elements only), and $\langle \delta(a-r_g)\rangle$ is the average accuracy in the batch. The factor $1/d_o$ reflects that $\mathcal{L}_{\text{Generator}}$ sums the divergence over observables, so the first term is an average over observables. All Experiment 2 results use $\mathcal{L}^{\star}_{\text{Generator}}$.  

\paragraph{Gradient stabilization (soft clip).}
To avoid vanishing gradients when clipping probabilities, we use a differentiable soft clip:
\[
\text{soft clip}(x, \epsilon_{clip}) = \epsilon_{clip} + \sigma\left(\frac{\hat{x}-\epsilon_{clip}}{\epsilon_{clip}}\right)(\hat{x}-\epsilon_{clip}),
\]
\[
\hat{x} = (1-\epsilon_{clip}) - \sigma\left(\frac{1-\epsilon_{clip}-x}{\epsilon_{clip}}\right)(1-\epsilon_{clip}-x).
\]
We found training reliability to improve dramatically when $\epsilon_{clip}=10^{-3}$. The soft clip is applied during training to the Classifier's marginal beliefs and to the probabilities entering each divergence.

\paragraph{Learnable Compositional Embeddings.}
Training reliability was improved when the learnable embedding estimates were over-parameterized. The relevant variable embedding estimates were passed through learnable linear layers $f_K, f_Q:\mathbb{R}^{d_{\hat{\mathcal{E}}}}\to\mathbb{R}^{d_{\mathcal{E}}}$, to project them down from $d_{\hat{\mathcal{E}}}=N$ to dimensionality $d_{\mathcal{E}}$. Estimated interactions were computed from the projections: 
$$\hat{\mathbf{z}}^i = \langle f_K(\hat{K}^i), f_Q(\hat{Q}^i) \rangle$$ 
Embedding regularization was applied to the projections:
$$\mathcal{L}_{\text{reg}} = \frac{1}{C\times d_o}\sum_{c,i}\bigl[(\| f_K(\hat{\mathbf{k}}^i_c)\|_2 - 1)^2 + (\|f_Q(\hat{\mathbf{q}}_c^i)\|_2 - 1)^2\bigr]$$

\begin{figure}[h!]
  \centering
 \includegraphics[width=1\columnwidth]{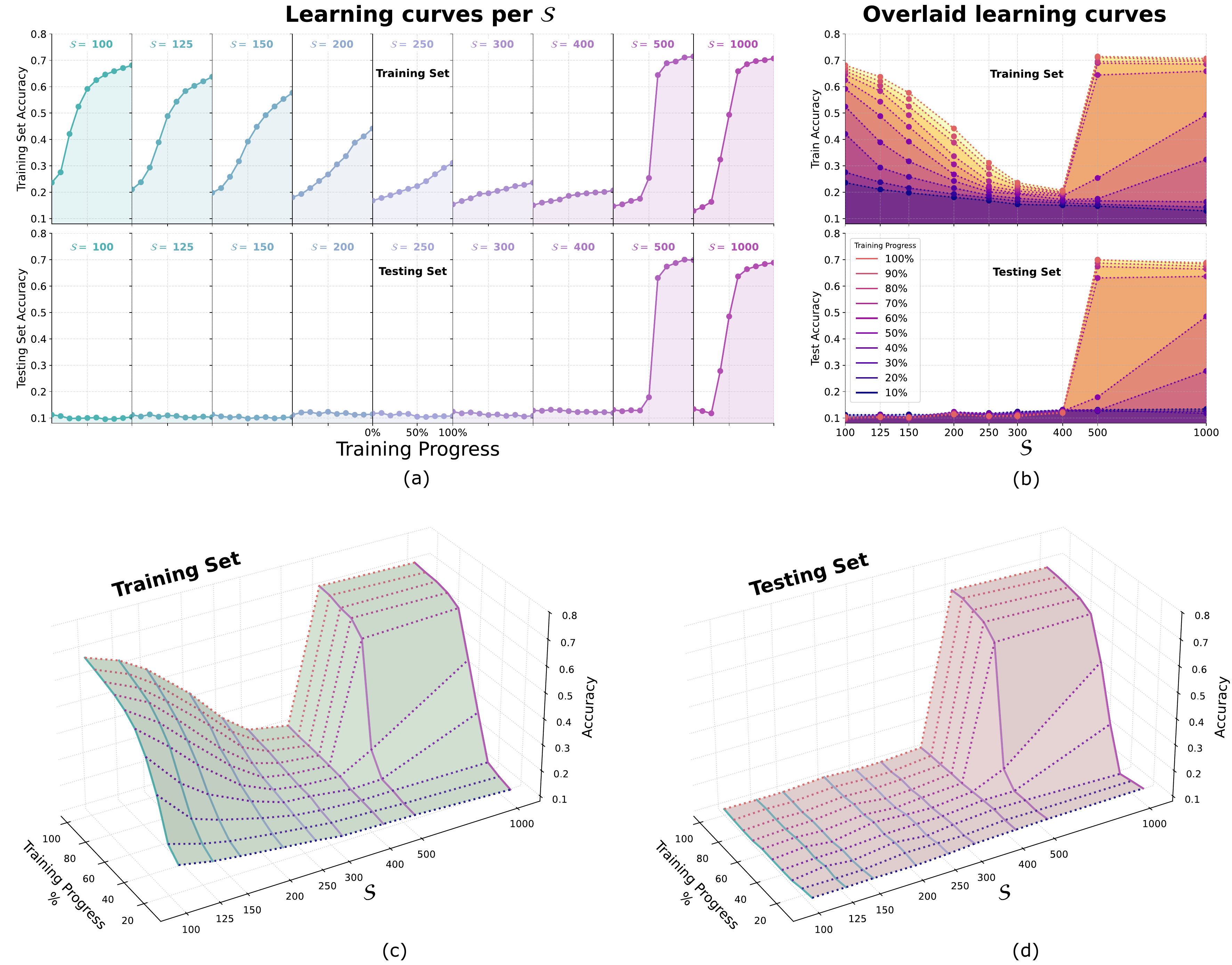}
  \caption{\textbf{Learning and generalization with Compositional Embeddings depend on $|\mathcal{S}|$ ($C =2$).} \textbf{(a)} Learning dynamics for training (top) and testing (bottom) variables, split by the number of latent variables $|\mathcal{S}|$. \textbf{(b)} All sizes of $\mathcal{S}$ overlaid and split by training progress. \textbf{(c-d)} 3-dimensional visualization of training and testing set accuracy as a function of $|\mathcal{S}|$ and training progress.}
  \label{fig:emb}
\end{figure}

\paragraph{Generalization emerges at a critical number of latent variables in set $\mathcal{S}$.} When $\hat{Z}$ was learned, we found that varying the total number of latent variables in the environment had a significant impact on generalization (\autoref{fig:emb}). When the environment contained too few latent variables ($|\mathcal{S}| < 500$), only training variables are learned. As the number of latent variables in $\mathcal{S}$ increased, learning on the training set slowed, until spontaneously recovering and generalizing to testing variables.

\begin{figure}[h!]
  \centering
 \includegraphics[width=1\columnwidth]{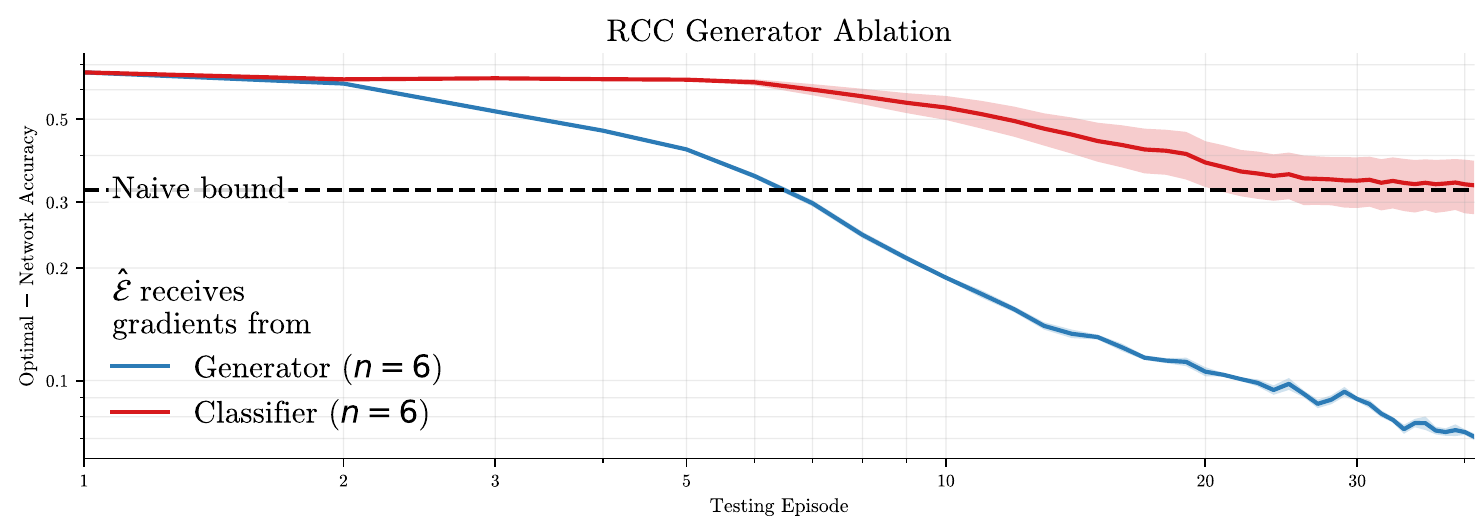}
  \caption{\textbf{Learning Compositional Embeddings through the classifier is unreliable ($C=2$)}. Excess error above optimality at the final step of inference (mean $\pm $ SEM with $6$ seeds), plotted against testing episode (1 test per 500 training episodes) on log–log axes. The embeddings receive gradients from either the Generator (blue) or Classifier (red).}
  \label{fig:ablation} 
\end{figure}

\paragraph{Generator ablation impairs performance.}
\label{app:ablation}
To isolate the contribution of the gradient from the Generator we trained RCCs in a simplified setting that removed all other challenges for learning. In both forward and backward passes, the Classifier's sampled actions were replaced by the ground truth realizations $\mathbf{r}$, providing an exact teaching signal to the Classifier and a reliable input to the Generator. Additionally, we used an orthonormal initialization of $f_K,f_Q$ with gain $\sqrt{N / d_{\mathcal{E}}}$ and zero bias to capture the statistics of $\mathcal{E}$. To further facilitate learning, we implemented layer normalization, skip connections and initialized the input layers of $\hat{Z}$ to be aligned with the readouts. \autoref{fig:ablation} demonstrates that RCC training is far more reliable when the gradient to $\hat{\mathcal{E}}$ flows from the Generator rather than from the Classifier.

\subsection{Experiment 3:}
\label{meth:exp3}

\paragraph{Controller loss.}

The Controller is trained using an Advantage Actor-Critic (A2C) algorithm to maximize the intrinsic reward:
\begin{equation} \label{eq:pref}
\mathcal{R}(\mathbf{r}, \mathbf{\Omega}, \ell) = \exp\left( \frac{1}{d_o}\sum_i \bigl(\ln(\ell_{\mathbf{z}^i}(\mathbf{r})) \Omega^i + \ln(1-\ell_{\mathbf{z}^i}(\mathbf{r}))(1-\Omega^i)\bigr) \right),
\end{equation}

The Actor and Critic both receive $\hat{Z}$, and output a policy over $\mathbf{r}$ and the expected $\mathcal{R}$, respectively. A single preferred observation $\mathbf{\Omega}$ is fixed for each Controller run, whereas the context, and hence $\hat{Z}$, varies across the elements of a batch; in the Offline conditions the same set of contexts is used in every episode. At each episode, $\mathbf{r}$ is sampled stochastically from the Actor's policy $\pi(\mathbf{r} \mid \hat{Z})$. In contrast to the Classifier, whose outputs are marginal beliefs, the Actor's policy is a probability distribution over the full action space so $\mathbf{r}$ is sampled from all dimensions jointly. The Critic predicts the intrinsic reward $\mathcal{R}$. The advantage $A$ is calculated as the difference between the true intrinsic reward and the Critic's prediction:
\[
A = \mathcal{R}(\mathbf{r}, \mathbf{\Omega}, \ell) - \nu(\hat{Z})
\]
The Controller loss $\mathcal{L}_{A2C}$ integrates the Critic's prediction error, the Actor's policy gradient and a policy entropy bonus weighted by $\beta_{cntrl}$ to encourage exploration:
\[
\mathcal{L}_{A2C} = A^2 - A_{\text{Detached}}\ln(\pi(\mathbf{r} \mid \hat{Z})) +\beta_{cntrl}\sum_{\mathbf{r}'}\pi(\mathbf{r}' \mid \hat{Z}) \ln \pi(\mathbf{r}' \mid \hat{Z}).
\]

\paragraph{Controller training comparisons.}
\label{app:offline_details}

\begin{figure}[h!]
  \centering
 \includegraphics[width=1\columnwidth]{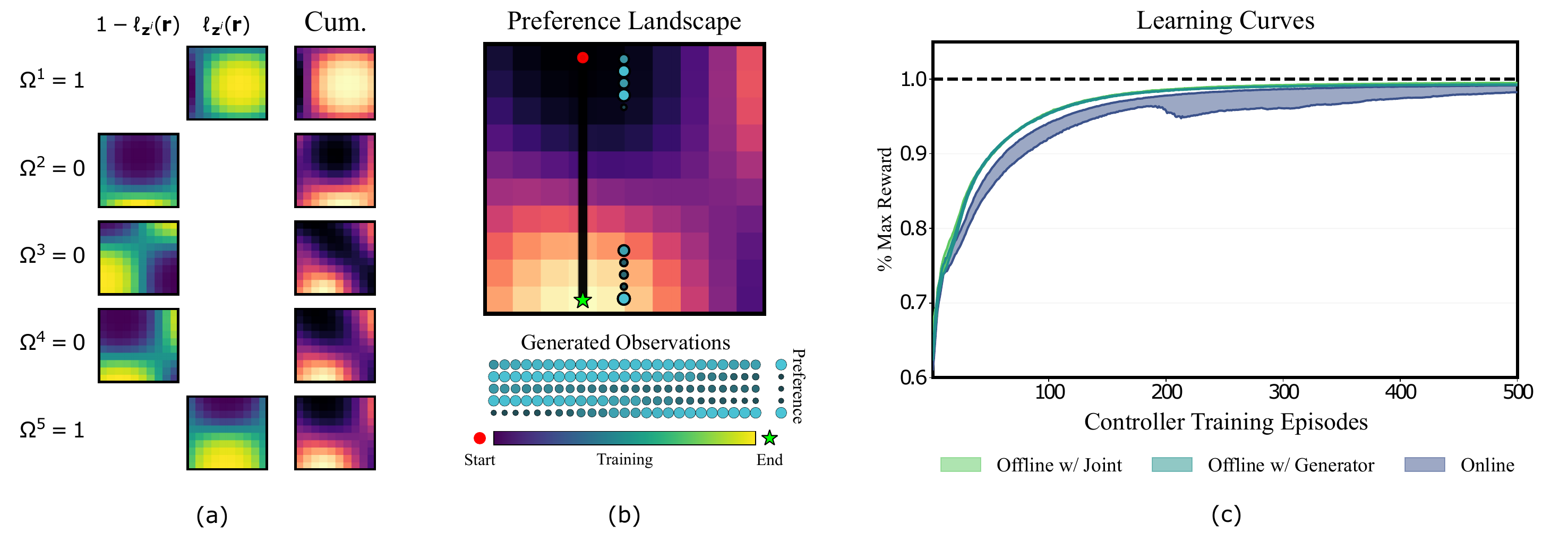}
  \caption{\textbf{Conditional generative modeling enables optimization in compositional spaces.} \textbf{(a)} Schematic illustrating the mapping of preferred observations ($\mathbf{\Omega}$) to their respective likelihoods and the cumulative landscape (accumulated over $i$ in \autoref{eq:pref}). \textbf{(b)} An example traversal, from the lowest to the highest point on the landscape, changes observations to best match the agent's preference. \textbf{(c)} Controller learning curves (mean $\pm$ SEM, $ n = 20$  initializations of $\mathbf{\Omega}$). $\%$ max reward is calculated as $\mathcal{R}(\arg\max_{\mathbf{r}}\pi(\mathbf{r} \mid \hat{Z}), \mathbf{\Omega}, \ell) \quad  / \quad \max_{\mathbf{r}}\mathcal{R}(\mathbf{r}, \mathbf{\Omega}, \ell)$, using the ground truth $\ell$.}
  \label{fig:RL}
\end{figure}

The schematic in \autoref{fig:RL}a demonstrates the mapping from preferences to specific likelihoods (and their cumulative product) into a \textit{preference landscape} of $\mathcal{R}(\mathbf{r}, \mathbf{\Omega}, \ell)$. \autoref{fig:RL}b shows mental navigation of the resultant preference landscape. We compare Controller training across three conditions that vary in how they estimate $\hat{Z}$ and $\ell$:

In the \textit{Online condition} \textbf{(i)}, the Controller directly interacts with the environment, setting the state of the world to a joint realization sampled from the Actor's policy. $\hat{Z}$ is learned from scratch by training a Generator (like in \autoref{meth:exp2}, but since this agent directly controls the world there is no uncertainty around $\mathbf{r}$ and therefore $\hat{\mathcal{E}}$ is learned much faster). Because every sampled $\mathbf{r}$ is correct, this Generator is trained without the reward weighting of $\mathcal{L}^{\star}_{\text{Generator}}$ (its divergence is averaged uniformly over the batch and observables), and its confidence input is fixed at 1. The likelihoods $\ell$ in $\mathcal{R}(\mathbf{r}, \mathbf{\Omega}, \ell)$ are approximated using the empirical average observation $\langle \mathbf{o} \rangle_t$ of the episode. 

The other two agents learn \textit{Offline}, sampling a joint realization $\mathbf{r}$ and imagining the resulting observations, without any manipulation of the environment. These agents receive an estimate of $\hat{Z}$ coming from embedding space $\hat{\mathcal{E}}$ which was passively learned through Representation Classification Chains (an RCC with the architecture and training procedure of \autoref{meth:exp2}, trained separately for this experiment with the hyperparameters listed in \autoref{app:ANE}). In order to approximate $\ell$ in $\mathcal{R}(\mathbf{r}, \mathbf{\Omega}, \ell)$, the \textit{Offline agent w/ Generator} \textbf{(ii)} uses the Generator that was pre-trained by the RCC to approximate $\ell_{\hat{\mathbf{z}}^i}(\mathbf{r}) \leftarrow \hat{P}_{\hat{\mathbf{z}}^i}(o^i=1\mid \mathbf{r})$. Serving as an oracle for comparison, the \textit{Offline agent w/ Joint Bayes} \textbf{(iii)} uses the ground-truth $\ell_{\mathbf{z}^i}(\mathbf{r}) \leftarrow P_{\mathbf{z}^i}(o^i=1\mid \mathbf{r})$. 

Training fully offline with the generative model achieves similar performance and efficiency to training offline with direct access to the joint likelihood (\autoref{fig:RL}c). Training online with direct interactions with the environment is no better than training offline. Training was performed with a batch of 20,000 trajectories per episode. In the Online condition, all 20,000 trajectories were followed by causal manipulations of the environment prior to the onset of the next episode.

\paragraph{Relation to offline model-based optimization.} Offline model-based optimization seeks to identify the input that maximizes an unknown objective given only a static dataset of inputs and their scores; a model of the objective is fit to these data and then optimized, which tends to produce inputs on which the model erroneously overestimates the objective \citep{Trabucco2021ConservativeOM}. Experiment 3 shares this structure, but the preference $\mathbf{\Omega}$ is internal and known, whereas the generative process $\ell$ had to be estimated and no objectively scored inputs exist. Optimizing against $\hat{P}_{\hat{Z}}(\mathbf{o}\mid\mathbf{r})$ exposes the Controller to the same failure mode as offline model-based optimization and policies trained inside learned world models \citep{Ha2018RecurrentWM}. Because all conditions are evaluated under the ground-truth $\ell$ (\autoref{fig:RL}c), such a failure would appear as a gap between the Offline agent w/ Generator and the oracle.

\paragraph{Choice of learner.} While Reinforcement Learning is not strictly necessary for solving this control problem (the joint action space $R^C=100$ was small enough to enumerate), our goal was to demonstrate that simple memory-less RL agents \textit{can} solve this problem. Modern memory-less RL algorithms could therefore be employed in more challenging settings where enumeration becomes impractical (as $C$ grows).

\section{SUPPLEMENTARY}

\paragraph{Important properties of likelihood constructions.}

We chose our manner of expansion to ensure several properties: 
\begin{itemize}
\item \textbf{Low rank structure:} Broadcasting $\mathbf{v}_{cc'}^i (\mathbf{v}_{c'c}^i)^\top \in \mathbb{R}^{R\times R}$ to $ \mathbb{R}^{{R}^{C}}$ constructs the likelihood entirely from pairwise interactions, embedding low-rank structure into the tensor.
\item \textbf{Constrained scale:} Using a sine function for $\mathbf{v}_{cc'}^i$ creates a linear (as a function of $C$) bound on the standard deviation of (pre-activation) elements of $\ell$.
\item \textbf{Visual Interpretability:} Constructing $\mathbf{v}^i_{cc'}$ with a smooth function ensures local similarity in $P_Z(\mathbf{o} \mid \mathbf{r})$ such that similar vectors $\mathbf{r}$ generate similar observations.
\item \textbf{Consistency:} Using $z_{cc'}^i$ to modulate phase, rather than scale, ensures different contexts and dimensions of $\mathcal{O}$ generate similar amounts of information.
\item \textbf{Useful Baseline:} Ensuring the potential for asymmetry in $\mathbf{v}^i_{cc'}$ ensures that naive inference can perform above chance.
\end{itemize}

\paragraph{Future Directions of the Generative Process.}
A natural extrapolation would be to construct $\mathbf{v}^i_{cc'}$ using Fourier series, perhaps adding another dimension to $K$ and $Q$ for frequency specific interaction. Such a design would embed structure at multiple spatial frequencies while staying interpretable, similar to modules of grid cells in the medial entorhinal cortex \citep{Burak2014SpatialCA}. Another natural extension would be to introduce sparsity to $K$ and $Q$, to move away from the current regime where all dimensions of $\mathcal{E}$ have \textit{some degree} of interaction. In summary, two promising future directions:
\begin{itemize}
    \item Construct $\ell$ through inverse discrete Fourier transform  (rather than a single frequency) for greater expressivity.
    \item Sparsify embeddings by masking elements in $\mathcal{E}$ for greater interaction specificity.
\end{itemize}

\subsection{Factorization Regret Relation to Interaction Information}
\label{app:sii_derivation}
For clarity we first focus on the Interaction Information between a single observable $o^i$ and two random variables, $I(r_g; r_c; o^i)$; the results extend to the full observation $\mathbf{o}$, whose channels are independent given $\mathbf{r}$. Since $r_g$ and $r_c$ are a priori independent in our setting, we can write:
$$I(r_g; r_c; o^i) = -I(r_g; r_c \mid o^i)= -\mathbb{E}[\mathcal{D}_{\mathrm{KL}}(P_{\mathbf{z}^i}(r_g, r_c \mid o^i) \,||\, P_{\mathbf{z}^i}(r_g \mid o^i) \otimes P_{\mathbf{z}^i}(r_c \mid o^i))]$$

Now consider two Bayesian observers processing a sequence of observations $o^i_{1:T}$:
\begin{enumerate}
    \item \textbf{Optimal Observer}: Updates its joint belief $B_t(r_g, r_c)$ correctly using $P_{\mathbf{z}^i}(o^i|r_g, r_c)$.
    \item \textbf{Naive Observer}: Updates its joint belief $\tilde{B}_t(r_g, r_c)$ using factorized likelihoods \[\tilde{P}_{\mathbf{z}^i}(o^i|r_g, r_c) \propto P_{\mathbf{z}^i}(o^i\mid r_g) \otimes P_{\mathbf{z}^i}(o^i\mid r_c).\]
\end{enumerate}

Since the Naive observer has separate likelihoods for $r_g$ and $r_c$, its marginal beliefs are independent:
\[
\tilde{B}_t(r_g \mid o^i_{1:t}) \;\propto\; p(r_g)\prod_{t'\leq t} P_{\mathbf{z}^i}(o^i_{t'} \mid r_g)\underbrace{\sum_{r_c} p(r_c)\prod_{t'\leq t} P_{\mathbf{z}^i}(o^i_{t'}\mid r_c)}_{\text{independent of } r_g},
\]
and it treats observations as i.i.d. given a single latent variable.

\paragraph{Interaction Information emerges in the joint posterior before impacting marginal beliefs.}
For a \textit{single observation ($o_1$)}, the Optimal observer's marginal belief: \[B_1(r_g \mid o^i_1) \propto p(r_g)\sum_{r_c}p(r_c)P_{\mathbf{z}^i}(o^i_1 \mid r_g, r_c)\] coincides exactly with the Naive observer's: \[
\tilde{B}_1(r_g \mid o^i_{1}) \;\propto\; p(r_g) P_{\mathbf{z}^i}(o^i_{1} \mid r_g).\] 
Yet a divergence can already exist between their joint posteriors, quantified by $I(r_g; r_c; o_1^i)$:
$$I(r_g; r_c; o^i_1) = - \mathbb{E} [\mathcal{D}_{\mathrm{KL}}(B_1(r_g, r_c|o^i_1) \,||\, \tilde{B}_1(r_g|o^i_1) \otimes \tilde{B}_1(r_c|o^i_1))]$$
Thus, for the first observation, the divergence between Optimal and Naive posteriors can contain Interaction Information, despite full agreement between their marginal beliefs. Critically, this Interaction Information plants the seed for the marginal beliefs to diverge on the next step of inference.

\paragraph{Factorization Regret measures the accumulating impact on marginal beliefs.}
The joint posteriors of the first step become the joint priors for the second. Any divergence between these joint priors allows \textit{subsequent observations} to create divergence in marginal beliefs. \textit{Factorization Regret (FR)} quantifies this accumulating divergence:
\[\mathbb{E}[\text{Factorization Regret}_T] = \mathbb{E} [ \mathcal{D}_{\mathrm{KL}}(B_T(r_g|\mathbf{o}_{1:T}) \,||\, \tilde{B}_T(r_g|\mathbf{o}_{1:T}))].\]
This consists of the conditional Total Correlation ($TC$) minus an offset. Applying the chain rule for $\mathcal{D}_{\mathrm{KL}}$ to the joint $P_Z(r_g, \mathbf{o}_{1:T})$ against its factorized counterpart $p(r_g)\prod_t P_Z(\mathbf{o}_t \mid r_g)$:
\[
\mathbb{E}[\text{Factorization Regret}_T] = \underbrace{\mathbb{E}\!\Big[\mathcal{D}_{\mathrm{KL}}\!\big(P_Z(\mathbf{o}_{1:T}|r_g)\;\|\;\textstyle\prod_t P_{Z}(\mathbf{o}_t|r_g)\big)\Big]}_{\;\defeq\; TC(\mathbf{o}_{1:T}\mid r_g)} \;-\; \mathcal{D}_{\mathrm{KL}}\!\big(P_Z(\mathbf{o}_{1:T}) \;\big\|\; \tilde{P}_Z(\mathbf{o}_{1:T})\big),
\]
where $\tilde{P}_Z(\mathbf{o}_{1:T}) = \sum_{r_g}p(r_g)\prod_t P_Z(\mathbf{o}_t \mid r_g)$. The term $TC(\mathbf{o}_{1:T}\mid r_g)$ expresses the reward-relevant information contained in the trajectory that is lost by a myopic focus on a goal.

\paragraph{Interactions in latent space generate temporal dependence between observations.}
The conditional total correlation $TC(\mathbf{o}_{1:T}\mid r_g)$ is a statement about \emph{time}, while $I(r_g;r_c;\mathbf{o})$ is a statement about \emph{latent space}. Because the realizations $\mathbf{r}$ are drawn once per episode and observations are i.i.d. given $\mathbf{r}$, the full set of latent variables explains all temporal dependencies:
\[ TC(\mathbf{o}_{1:T}\mid r_g) \;=\; \underbrace{TC(\mathbf{o}_{1:T}\mid r_g, r_c)}_{=\;0} \;+\; \sum_t I(\mathbf{o}_t; r_c \mid r_g) \;-\; I(\mathbf{o}_{1:T}; r_c \mid r_g)\]
 Because sequential inference is permutation invariant we can consider a generic observation $\mathbf{o}$:
\[ TC(\mathbf{o}_{1:T}\mid r_g) \;=\; T\, \times ( I(\mathbf{o}; r_c) - I(\mathbf{o}; r_c; r_g)) \;-\; I(\mathbf{o}_{1:T}; r_c \mid r_g).\]
The first term is positive because $I(\mathbf{o};r_c;r_g)\leq0$ and the second term is bounded by the entropy of $r_c$. Thus, the total correlation may grow linearly with time, with Interaction Information $I(\mathbf{o};r_c;r_g)\leq0$ determining the average rate of growth. This reveals a direct mapping from the geometry in latent space to the temporal dependencies in a trajectory.

\paragraph{Relation to Semantic Information.} \citet{Kolchinsky2018SemanticIA} define semantic information as the information a system has about its environment that is causally necessary for maintaining its \textit{viability} (i.e. existence), and identify it through interventions that scramble system--environment correlations. The Naive observer applies such an intervention to Bayesian inference: it replaces the joint likelihood with the product of its marginals, scrambling the Interaction Information while preserving each variable's marginal likelihood. Factorization Regret measures the consequence of this intervention for the goal, with the belief over the optimal action in the role of viability. It is therefore the analogue of their value of information under full scrambling. The single-observation case above makes the distinction concrete: Interaction Information is already present in the joint posterior, yet Factorization Regret is zero because the belief over the goal is unaffected. Interaction Information thus becomes semantic only once it bears on the goal.

\subsection{Learning Dynamics}
\label{app:B2}

\begin{figure}[h!]
  \centering
 \includegraphics[width=1\columnwidth]{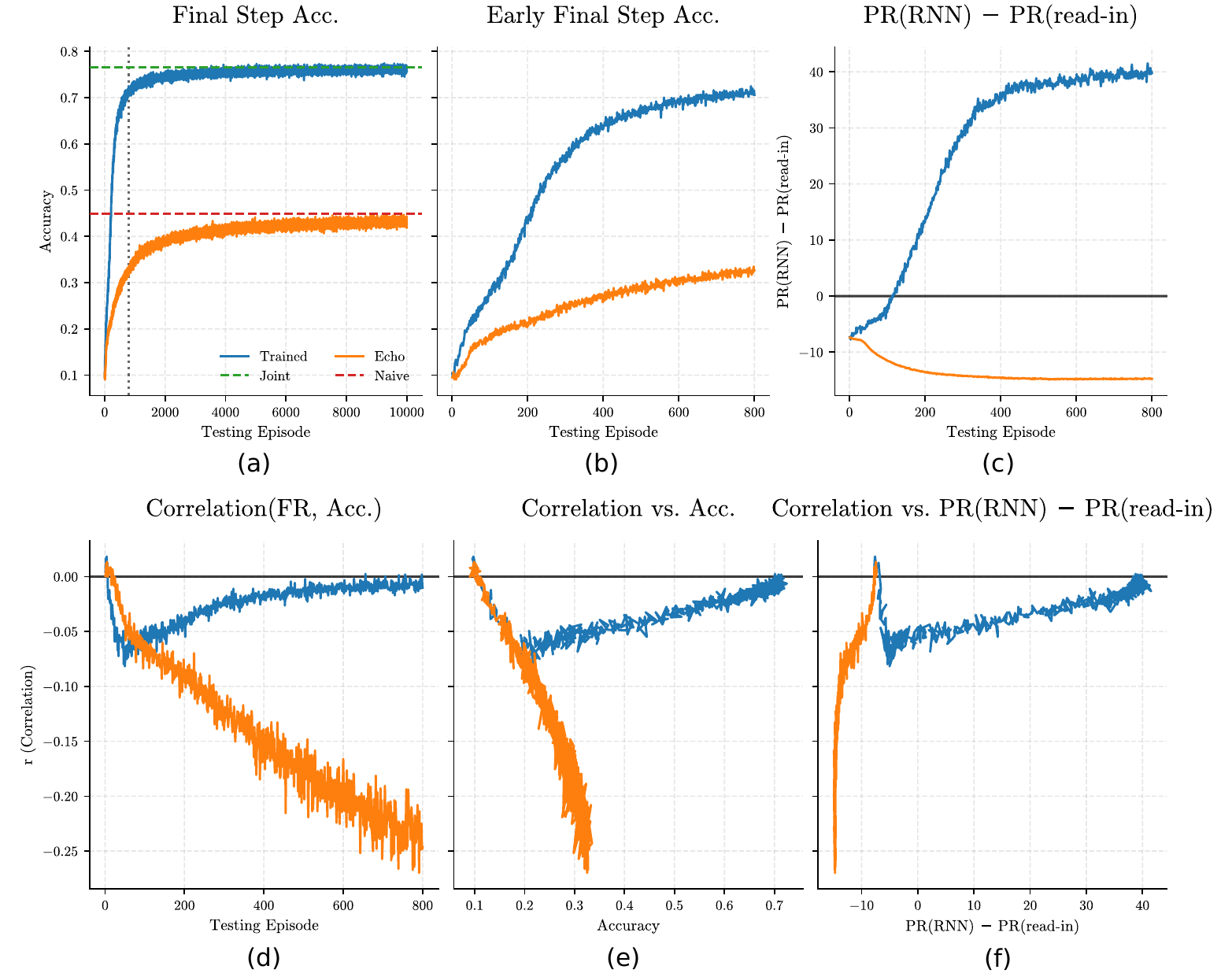}
\caption{\textbf{Managing Factorization Regret involves high-dimensional RNN dynamics ($C=2$).}  \textbf{(a)} Throughout learning, testing accuracy at the final step of inference saturates to the performance of either Joint or Naive Bayes. The vertical dotted line marks the range shown in (b--f). \textbf{(b)} Final step testing accuracy during early training.  \textbf{(c)} Difference in Participation Ratio (PR) between the recurrent and read-in activity. \textbf{(d)} Correlation between Factorization Regret and accuracy at the final step of inference. A negative correlation emerges in the beginning of training and persists for the ESN. \textbf{(e)} Correlation versus accuracy. \textbf{(f)} Correlation versus PR difference.}
\label{fig:lrning}
\end{figure}

 Average performance improves throughout training for both Fully Trained RNNs and ESNs (\autoref{fig:lrning}a-b). However, we hypothesize that the ESN is immediately on a trajectory towards a suboptimal \textit{stimulus-bound} strategy \citep{Vertechi2019InferenceBasedDI}. To test this hypothesis, we investigate the Participation Ratio (i.e. effective dimensionality) of the Read-in and recurrent activity (\autoref{fig:lrning}c). While a Fully Trained network expands the dimensionality of its recurrent dynamics relative to the Read-in, the ESN shows the opposite effect. Although the ESN accuracy improves, a large input gain, relative to the recurrent dynamics, could make it difficult to learn when subtle changes in a trajectory correspond to large jumps in latent space. 
 
 We next investigated whether Factorization Regret could reveal the strategies being learned. Early into training a negative correlation emerges between accuracy and Factorization Regret across the batch of trajectories (\autoref{fig:lrning}d). Since the networks initially know nothing, the correlation between FR and accuracy begins at zero. Failure rapidly becomes predictable from FR; however, this correlation dissolves for the Fully Trained RNN and persists in the ESN. To demonstrate that this is not a trivial outcome of performance differences, we plot this correlation against the average accuracy in the episode (\autoref{fig:lrning}e), demonstrating that even when networks have the same accuracy on average, they are developing different strategies with respect to FR. This capacity to mitigate the correlation between FR and accuracy appears to coincide with an increase in recurrent dimensionality on average (\autoref{fig:lrning}f).

\subsection{Entanglement} \label{app:B3}

\begin{figure}[h!]
  \centering
  \includegraphics[width=0.48\linewidth]{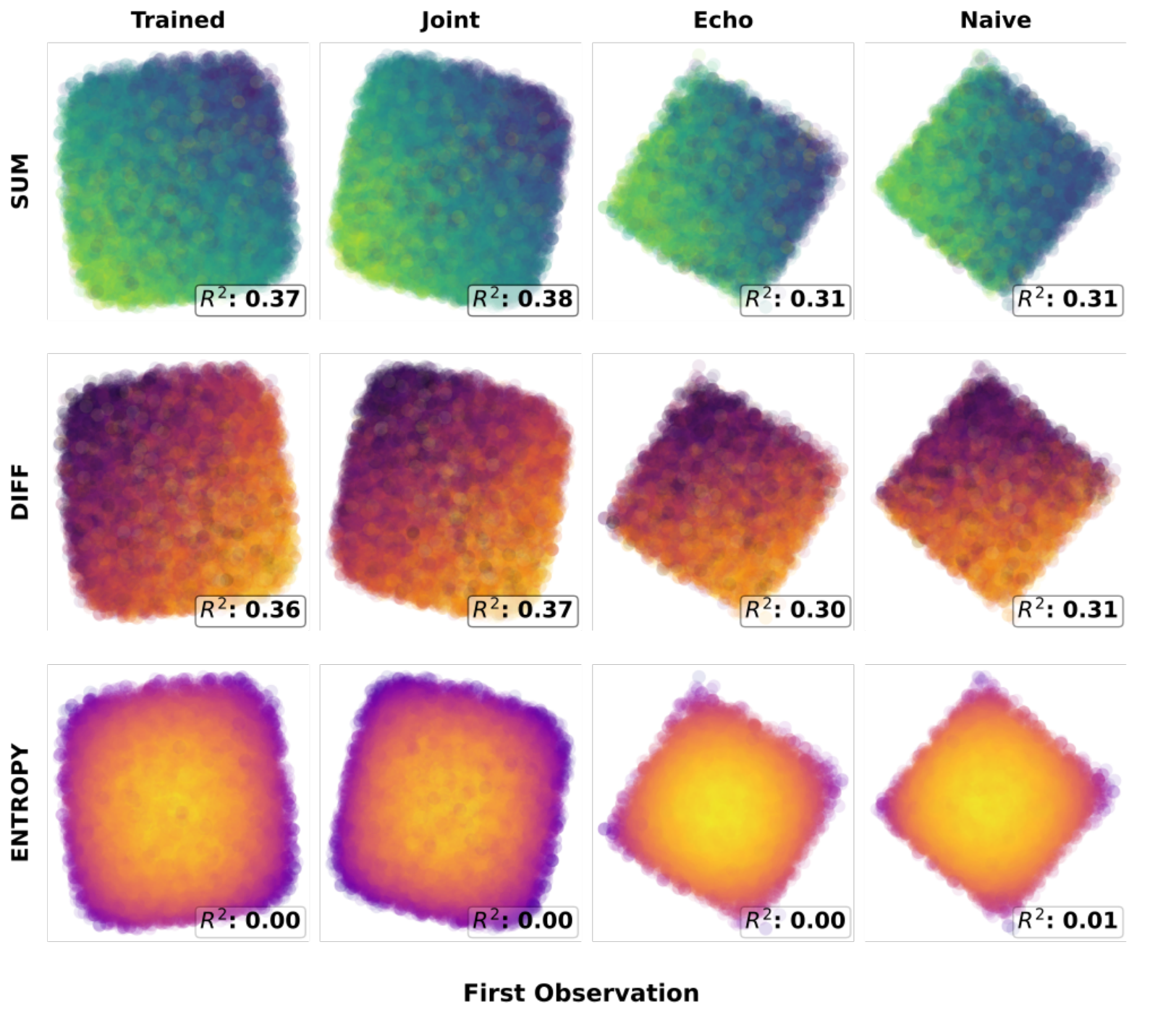}
  \hfill
  \includegraphics[width=0.48\linewidth]{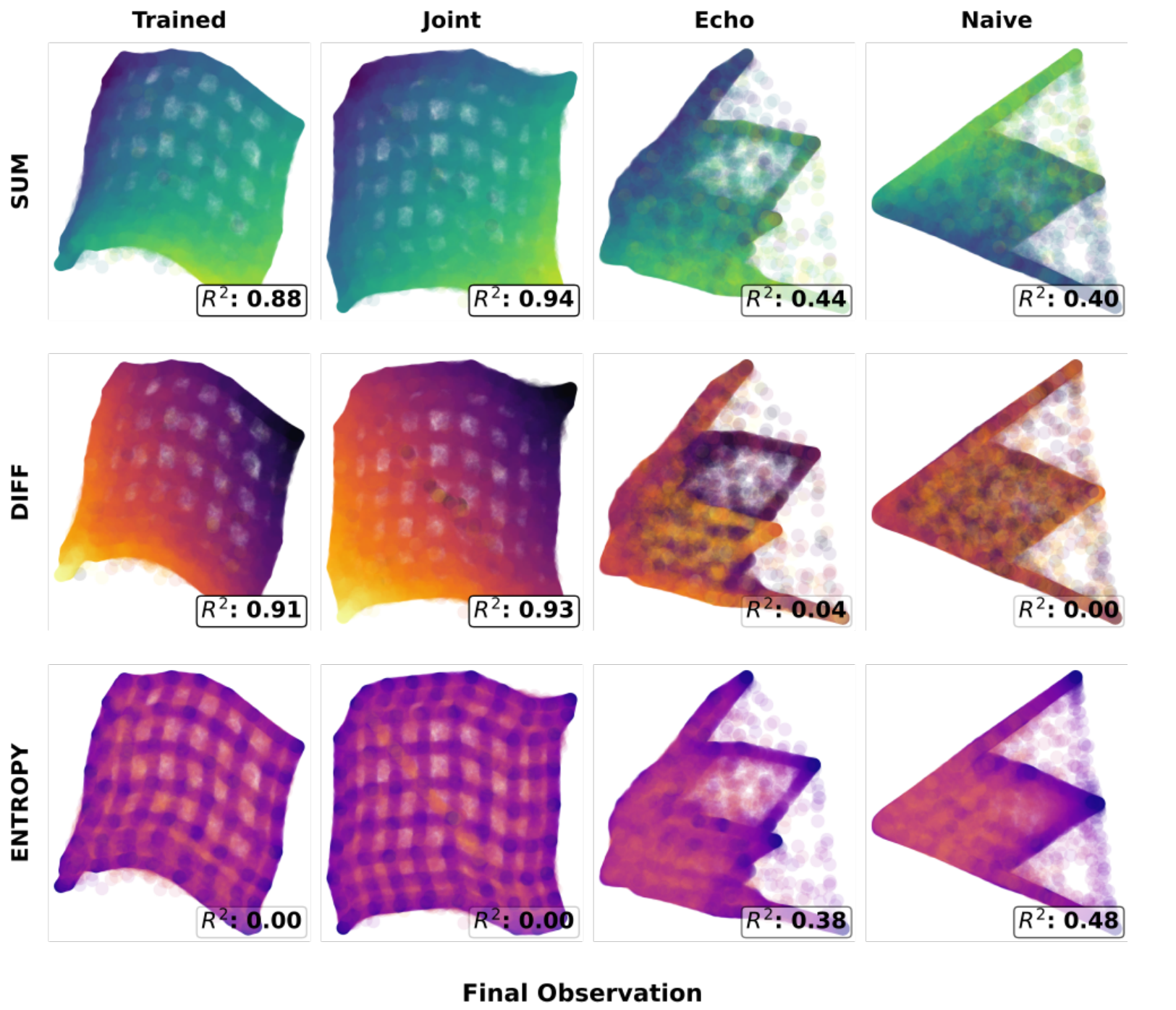}
  \caption{\textbf{Factorization Regret leads to entanglement over time.} Top 2 Principal Components of the marginal beliefs after only a single observation (left) and $T = 30$ observations (right). Here $c$ and $c'$ denote the two variables in context ($C=2$), one of which is the goal. Beliefs are colored by the realization sum ($r_c + r_{c'}$), difference ($r_c - r_{c'}$), and belief entropy ($-\sum_c\sum_r B_{tcr }\ln B_{tcr}$). $R^2$ indicates the variance explained of the respective variables by the top 2 components. After only a single observation, representations are still perfectly factorized. By the final observation, ESN and Naive representations become deeply entangled, while the Fully Trained representation remains factorized alongside Optimal Joint Bayes.}
\label{fig:mani}
\end{figure}

Misalignment between the ESN's internal space and the true latent space could be explained mechanistically as entanglement. Entanglement in an RNN's representation may emerge from the mixing of information by temporal correlations between neurons \citep{miller2023cognitive}.  Therefore, we hypothesized that training only a read-in and read-out on an evidence accumulation task, which demands temporal correlations, would not only fail to overcome this entanglement, but would exacerbate it. Indeed, Principal Component Analysis (PCA) on the marginal beliefs for network and Bayesian agents was consistent with this prediction (\autoref{fig:mani}). When performing PCA on marginal beliefs after a single observation, all representations appeared factorized. Yet PCA on the marginal beliefs at the end of inference appeared entangled for the ESN and Naive Bayes which lose the ability to differentiate $r_c$ from $r_{c'}$. Additionally, the Joint and Fully Trained representations allocate low entropy beliefs (i.e. high confidence) to all $(r_c, r_{c'})$ pairs. In contrast, the ESN and Naive Bayes represent confidence orthogonally to $r_c$ and $r_{c'}$.

% \begin{figure}[H]
%   \centering
%  \includegraphics[width=.9\columnwidth]{figures/PDFs/pca_manifolds_first3.pdf}
% \caption{\textbf{Belief representations are initially factorized.} Top 2 Principal Components of the marginal beliefs after a single observation. Beliefs are colored by the realization sum ($r_c + r_{c'}$), difference ($r_c - r_{c'}$), and belief entropy ($-\sum_c\sum_r B_{tcr }\ln B_{tcr}$). $R^2$ indicates the variance explained of the respective variables by the top 2 components.}
% \label{fig:mania}
% \end{figure}

% \begin{figure}[H]
%   \centering
%  \includegraphics[width=.9\columnwidth]{figures/PDFs/pca_manifolds_final3.pdf}
% \caption{\textbf{Failing to capture interaction information creates entanglement.} Same as \autoref{fig:mania}, for the final observation. The Fully Trained network maintains factorized representations akin to Joint Bayes. The Echo State and Naive representations of $r_c$ and $r_{c'}$ become entangled.}
% \label{fig:manib}
% \end{figure}

For a mechanistic account of this entanglement, let $\mathbf{e}_r \in \{0,1\}^R$ represent a one-hot indicator vector which inquires about the probability of realization $r$. In $P_{z^i}(o^i \mid r)$, the mapping from scalar $z^i$ to a vector of $R$ updates is mediated by the pre-sigmoid expansion $f(z^i)$ (the vector $\mathbf{v}^i$ of \autoref{app:de}). For $C=1$, inquiring about a specific realization $r=r'$ projects the indicator vector onto the likelihood via:
\[
P_{z^i}(o^i = 1 \mid r=r') =   \sigma\bigl(\mathbf{e}_{r'}^Tf(z^i)\bigr).
\]

Crucially, because \textbf{different realizations of the same variable are mutually exclusive}, the likelihood of a specific $r = r'$ is perfectly isolated from all other potential realizations and $P_{Z}(r=r' \mid \mathbf{o})$ may be updated independently. Next, consider a context of two latent variables ($C=2$). Here the interactions $Z^i = (z^i_{gc}, z^i_{cg})$ map to a joint space via the bilinear tensor product $f(z^i_{gc}) \otimes f(z^i_{cg}) \in \mathbb{R}^{R \times R}$, forming an $R \times R$ matrix of belief updates. The bilinear construction generates synergy. In other words, the combination of variables loses predictive power if the variables are considered in isolation, i.e. the whole is greater than the sum of its parts \citep{Proca2022SynergisticIS, Barak2013TheSO}. We can see this because evaluating the likelihood of $o^i=1$ given a realization $r_g = r'$ now involves projecting indicator $\mathbf{e}_{r_c}$ onto a basis shared with $\mathbf{e}_{r_g}$:
\[
P_{Z^i}(o^i = 1 \mid r_g=r', r_c) = \sigma\Bigl(\mathbf{e}_{r'}^T \left( f(z^i_{gc}) f(z^i_{cg})^T \right) \mathbf{e}_{r_c}\Bigr).
\]
In summary, the entanglement between latent variables emerges from projecting onto a shared basis.

Next we turn to investigating entanglement in real-time. Since FR is an accumulating measure (best used for comparing full trajectories) it is not well suited for high temporal fidelity analysis. Instead, we aim to compare the Naive marginal likelihood $P_Z(\mathbf{o}_t \mid r_g)$ and the history-conditioned marginal likelihood $P_Z(\mathbf{o}_t \mid r_g, \mathbf{o}_{1:t-1})$. The history-conditioned marginal likelihood is proportional in $r_g$ to the ratio of the Optimal Observer's successive marginalized beliefs, since the evidence term $P_Z(\mathbf{o}_t \mid \mathbf{o}_{1:t-1})$ does not depend on $r_g$: $$P_Z(\mathbf{o}_t \mid r_g, \mathbf{o}_{1:t-1}) = \sum_{r_c}P_Z(\mathbf{o}_t \mid r_g, r_c)p(r_c \mid r_g, \mathbf{o}_{1:t-1})\propto \frac{B_{t}(r_g)}{B_{t-1}(r_g)}.$$ 

The history-dependence of the Optimal Bayesian update can be attributed to the prior's impact on the functional likelihood, $ \sum_{r_c}P_Z(\mathbf{o}_t \mid r_g, r_c)p(r_c \mid r_g, \mathbf{o}_{1:t-1})$, manifesting as entanglement between variables $r_g$ and $r_c$. To quantify this \textit{``non-Markovian-ness''}, we measure the Jeffreys Divergence between (normalized) Naive marginal likelihoods and history-conditioned marginal likelihoods:
$$\text{Entanglement}_t =  J\bigl( \frac{P_Z(\mathbf{o}_t \mid r_g)}{\eta_a} \;||\;\frac{P_Z(\mathbf{o}_t \mid r_g, \mathbf{o}_{1:t-1})}{\eta_b} \bigr) = \! J\bigl( \frac{P_Z(\mathbf{o}_t \mid r_g)}{\eta_a} \;||\;\frac{B_{t}(r_g)}{B_{t-1}(r_g)\, \eta_b} \bigr)\!, $$
where $\eta_a$ and $\eta_b$ are normalization factors over $r_g$ to ensure valid probabilities; normalization removes the $r_g$-independent evidence term, so the two expressions are equal.

\begin{figure}[h!]
  \centering
 \includegraphics[width=1\columnwidth]{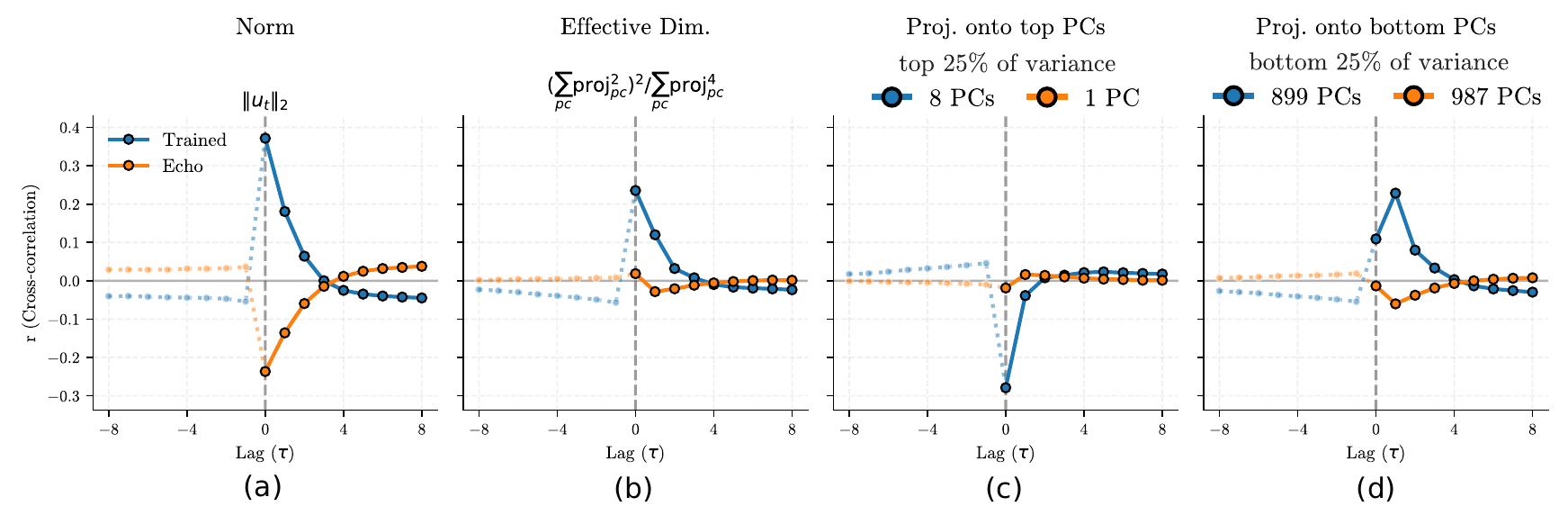}
\caption{\textbf{Entanglement correlates with learned dynamics of dimensionality.}  Cross-correlations between Bayesian entanglement and \textbf{(a)} the L2 norm of RNN output $u_t$, \textbf{(b)} the effective dimensionality of the projection onto the Principal Components, \textbf{(c)} the portion of the effective dimensionality explained by activity in the band of top Principal Components that explains $25\%$ of neural variance and \textbf{(d)} same as (c) for the band of bottom Principal Components. Only data from $t >20$ was analyzed in order to isolate the steady-state response profile. Legends give the number of Principal Components in each band. }
\label{fig:crss}
\end{figure}

To investigate the response of the network dynamics to entanglement, we performed a cross correlation between Optimal Bayesian belief $\text{Entanglement}_t$ and various properties of the network output. \autoref{fig:crss}a demonstrates that entanglement is associated with a change of activity in population space, in both Fully Trained RNNs and ESNs. However, the response differs greatly between networks. In the Fully Trained RNN, Entanglement is associated with an expansion of the population vector, whereas the ESN activity appears to shrink towards the origin. The Fully Trained RNN's expansion is also distributed throughout the population, rather than along the top component directions (\autoref{fig:crss}b). Interestingly, the top components of the Fully Trained RNN's activity are suppressed (\autoref{fig:crss}c), and the expansion is unexpectedly spread across many low variance components (\autoref{fig:crss}d). These dynamics could account for the increased RNN dimensionality observed in \autoref{fig:lrning}c, indicating that dynamic control of network dimensionality may be a mechanism for maintaining a disentangled representation.

\subsection{Representation Classification Chains and Thousand Brains Theory}
The Thousand Brains Theory of intelligence \citep{Hawkins2018AFF} postulates that individual cortical columns autonomously model through \textit{compositions of compositions}. Tasked specifically with learning useful relationships, Representation Classification Chains can be viewed as the potential \textit{microcircuit} of such a system. This architecture-agnostic circuit generates abstractions $\mathbf{r}$, to explain inputs $\mathbf{o}$, in a manner that enables optimal control of inputs given their internal value.

\subsection{Additional Model Hyperparameters}
\label{app:ANE}

All models are implemented in PyTorch.

\paragraph{General.}
\begin{itemize}
 \item Hidden dimensionality of all networks ($N$): 1000
  \item Parametric embedding dimensionality ($d_{\hat{\mathcal{E}}}$): 1000
  \item Optimizer: Adam
\end{itemize}

\paragraph{Experiment 1.}
\begin{itemize}
  \item Learning rate: 0.001
  \item Training episodes: 50000
  \item Training batch size: 8000
  \item Loss exponent ($\alpha$): $0.5$
\end{itemize}

\paragraph{Experiment 2.}
\begin{itemize}
  \item RCC Learning rate: 0.0005
  \item RCC Training episodes: 125000
  \item RCC Training batch size: 8000
  \item RCC Regularization weight ($\beta_{1}$): $0.1$
  \item Testing: once per 100 training episodes

\end{itemize}

\paragraph{Experiment 3.}
\begin{itemize}
  \item RCC Learning rate: 0.0001
  \item RCC training episodes: 50000
  \item RCC Training batch size: 20000
  \item RCC Regularization weight ($\beta_{1}$): $0.01$
  \item Controller Learning rate: 0.005
  \item Controller Training episodes: 500
  \item Controller Batch size: 20000
  \item Controller Entropy bonus ($\beta_{cntrl}$): $0.05$
  \item Online Generator Learning Rate: 0.001
\end{itemize}

\textbf{Classifier Architecture:} \newline
\begin{enumerate}
    \item \textbf{Input:} At each time step $t$, the input $x_t^{cls}$ is the concatenation of the current observation vector $o_t$ and the interactions $Z$ (or their estimate $\hat{Z}$ in Experiments 2--3): $x_t^{cls} = [o_t, Z]$.
    
        \item \textbf{Read-in:} $h_t^{in} = \text{ReLU}(W_{in}x_t^{cls} + b_{in})$ maps the input to the hidden dimension $N$.
        \item \textbf{LSTM:} $h_t^{stm}, h_t^{ltm} = \text{LSTM}(h_t^{in}, (h_{t-1}^{stm}, h_{t-1}^{ltm}))$ processes the $T\times N$ sequence where initial hidden states $(h_0^{stm}, h_0^{ltm})$ are learnable parameters.
        \item \textbf{Read-out:} $l_t = W_{out}h_t^{stm} + b_{out}$ maps the LSTM output to $C$ independent sets of $R$ dimensional logits.
    \item \textbf{Output:} As described in the main text, a cumulative sum is applied along time and then a $\text{softmax}$ is applied along the realizations to produce the Classifier's belief-state.
\end{enumerate}
The ESN uses the same architecture with its recurrent weights frozen at PyTorch's default initialization, with the read-in and read-out trained. Its reservoir therefore has $N=1000$ units, with no spectral-radius rescaling or sparsification.

\textbf{Generator Architecture:} \newline 

 \textbf{Input:} The Generator integrates three inputs: $x^{gen} = h_r + h_c + h_z, \quad \text{where } h_r,h_c, h_z \in \mathbb{R}^{N}.$
\begin{itemize}
    \item $h_r = W_s\, \mathrm{Emb}(\mathbf{\hat{r}}) + b_s $ is a linear projection of learnable PyTorch embeddings $\mathrm{Emb}(\mathbf{\hat{r}})$, concatenated across the $C$ variables in context.
    \item $h_c = W_c \hat{\mathbf{r}}_{conf} + b_c $ is a linear projection of the confidence in the sampled realizations.
    \item $h_z = W_z \hat{\mathbf{z}}^i + b_z $ is a linear projection of estimated interactions $ \hat{\mathbf{z}}^i$. Since dimensions of the observation space are independent we process them in parallel.
\end{itemize}
    \textbf{Output:} $\hat{P}_{\hat{Z}}(\mathbf{o}|\hat{\mathbf{r}}) = \sigma \bigl(W_2( \text{ReLU}(W_1(\text{ReLU}(x^{gen})) + b_1)) + b_2\bigr).$

\textbf{Controller Architecture:}\newline 

The Actor and Critic networks receive only $\hat{Z}$, and pass it through separate two-layer MLPs with ReLU activations. For the Actor, the final output is passed through a $\text{Softmax}$ to produce a distribution over actions. For the Critic, the output is wrapped in a $\text{Sigmoid}$ to predict the reward.

\subsection{Pseudo-code}
\begin{algorithm}[H]
   \caption{Cognitive Gridworld Task}
   \label{alg:generative_process}
\begin{algorithmic}
   \STATE {\bfseries Input:} Total latent variables $|\mathcal{S}|$, Context size $C$, Realizations $R$, Time Horizon $T$
   \STATE {\bfseries Hyperparameters:} Temperature $\lambda$
   \STATE {\bfseries Initialize Embeddings:} $K, Q \leftarrow$ \textsc{Create Compositional Embeddings}($|\mathcal{S}|$)

   \STATE
   \STATE {\bfseries Create Compositional Embeddings:}
       \STATE $K, Q \sim \mathcal{N}(0, I)$ \hfill \COMMENT{Draw matrix of size $|\mathcal{S}| \times d_o \times d_{\mathcal{E}}$}
       \FOR{$s \in \{1, \dots, |\mathcal{S}|\}$}
            \STATE $K_s \leftarrow \textsc{GramSchmidt}(K_s)$ \hfill \COMMENT{Orthogonalize across $d_o$}
            \STATE $Q_s \leftarrow \textsc{GramSchmidt}(Q_s)$
       \ENDFOR
       \STATE $K \leftarrow K / ||K||_2, \quad Q \leftarrow Q / ||Q||_2$ \hfill \COMMENT{Normalize across $d_{\mathcal{E}}$}
       \RETURN $K, Q$

   \STATE
   \STATE {\bfseries Compression \& Expansion(K,Q):}
   \STATE $L \leftarrow 2R+1$ \hfill \COMMENT{Profile length}
   \STATE Initialize sinusoidal base $\mathrm{base}(n)$ of one period over $L$ slots, scaled by $\lambda$
      \FOR{ordered pair $c, c' \in \mathcal{C}$, $c \neq c'$ (or $c = c'$ if $C = 1$)}      
    \STATE $z \leftarrow \langle \mathbf{k}^i_c, \mathbf{q}^i_{c'} \rangle$ \hfill \COMMENT{Get cosine similarity}
    \STATE $d(n) \leftarrow$ circular distance from $n$ to $L z$ on $L$ slots
    \STATE $\omega \leftarrow \mathrm{softmax}_n(-d(n)^2)$ \hfill \COMMENT{Smoothing kernel centered at $Lz$}
    \STATE $v_{cc'}^i(r) \leftarrow \sum_n \omega(n)\,\mathrm{base}((n-r) \bmod L)$ \hfill \COMMENT{Smooth roll of base}
    \ENDFOR

   \IF{$C = 1$}
   \STATE $\ell_{\mathbf{z}^i} \leftarrow \sigma(v_{cc}^i)$
   \ELSE
   \STATE $\ell_{\mathbf{z}^i} \leftarrow \sigma\biggl(\sum_{c<c'} \text{Broadcast}(\{v_{cc'}^i (v_{c'c}^i)^\top\})\biggr)$ \hfill \COMMENT{Tensor of shape $R^{C}$ parameterized by $Z$}
   \ENDIF

   \RETURN $\ell_{\mathbf{z}^i}$ \hfill \COMMENT{Likelihood of observing $'1'$ in dimension $i$ of $\mathbf{o}$}

   \STATE
   \STATE {\bfseries Episode Generation:}
   \STATE Sample context indices $\mathcal{C} \subset \{1, \dots, |\mathcal{S}|\}$, $|\mathcal{C}| = C$
   \STATE Sample realizations $r_c \sim \text{Uniform}(R)$ for $c \in \mathcal{C}$
   \STATE Sample goal $g \sim \text{Uniform}(\mathcal{C})$
   \STATE $a^* \leftarrow r_g$
   \FOR{observation dimension $i \in \mathcal{O}$}
           \STATE $\ell_{\mathbf{z}^i} \leftarrow \text{Compression \& Expansion}(K^i,Q^i)$
   \ENDFOR

   \STATE Generate trajectory $\mathbf{o}_{1:T}$ where $o_t^i \sim \text{Bernoulli}(\ell_{\mathbf{z}^i}(\mathbf{r}))$
\end{algorithmic}
\end{algorithm}

%%%%%%%%%%%%%%%%%%%%%%%%%%%%%%%%%

\begin{algorithm}[H]
   \caption{Experiment 1: Training to capture Factorization Regret}
   \label{alg:experiment_1}
\begin{algorithmic}
   \STATE {\bfseries Input:} Episodes $E$, Interactions $Z$,  \textsc{FullyTrained} or \textsc{EchoState} Network $\Phi$       

\STATE 
\FOR{each episode $(\mathbf{o}_{1:T}, r_g, Z) \in E$}
\STATE $h_0 \leftarrow h_{\phi}$ \hfill \COMMENT{Initialize memory streams}
\STATE $\mathcal{L} \leftarrow 0$ \hfill \COMMENT{Initialize loss}
\STATE $\ln B^{\text{Net}}_0 \leftarrow \mathbf{0}$ \hfill \COMMENT{Initialize log belief}
\STATE $\ln B^{\text{Joint}}_0 \leftarrow \mathbf{0}$  \hfill \COMMENT{Initialize log belief}
\FOR{$t=1$ {\bfseries to} $T$}
       \STATE $[h_{t},\Delta] \leftarrow \Phi([\mathbf{o}_t, Z], h_{t-1})$
       \STATE $\ln B_{t}^{\text{Net}}  \leftarrow \ln B_{t-1}^{\text{Net}} +  \Delta $ 
       \STATE $ \ln B_{t}^{\mathrm{Joint}}(\mathbf{r}) \leftarrow \ln B_{t-1}^{\mathrm{Joint}}(\mathbf{r}) + \ln P_Z(\mathbf{o}_t|\mathbf{r}) - \ln P_Z(\mathbf{o}_t)$ \hfill \COMMENT{For all $\mathbf{r}$}
       \STATE $B_{tg}^{\mathrm{Joint}} \leftarrow \sum_{\mathbf{r}_{\setminus g}} \text{Softmax}(\ln B_{t}^{\mathrm{Joint}})$ \hfill \COMMENT{Goal marginal}
       \STATE $\mathcal{L} \leftarrow \mathcal{L} + \frac{1}{T} \mathcal{L}_{\text{FR}}$ \hfill \COMMENT{Using $B_{t}$, see \autoref{meth:exp1}}
       
   \ENDFOR
   \STATE Update $\Phi, h_{\phi}$ to minimize $\mathcal{L}$
   \ENDFOR

\end{algorithmic}
\end{algorithm}

%%%%%%%%%%%%%%%%%%%%%%%%%%%%%%%%%%%%%%%%%%%%%%%%%%%%%%%%%%%%%%%%%%%%%%

\begin{algorithm}[H]
   \caption{Experiment 2: Training Representation Classification Chains}
   \label{alg:experiment_2}
\begin{algorithmic}
   \STATE {\bfseries Input:} Episodes $E$, Classifier $\Phi$, Generator $\Psi$
   \STATE {\bfseries Hyperparameters:} Regularization weight $\beta$
   \STATE {\bfseries Learnable Embeddings:} $\hat{K}, \hat{Q}$, $f_K, f_Q$
   
   \STATE 
   \FOR{each episode $(\mathbf{o}_{1:T}, r_g, \mathcal{C}) \in E$}
      \STATE $\hat{Z} \leftarrow \langle f_K(\hat{K}_{\mathcal{C}}), f_Q(\hat{Q}_{\mathcal{C}}) \rangle$ \hfill \COMMENT{Estimated Interactions}
    \STATE $ \ln B^{\text{Net}}_0 \leftarrow \mathbf{0} $
      \STATE $h_0 \leftarrow h_{\phi}$

       \STATE 
     \STATE {\bfseries $\Phi$ Forward Pass:}
      \FOR{$t=1$ {\bfseries to} $T$}
         \STATE $[h_{t},\Delta] \leftarrow \Phi([\mathbf{o}_t, \hat{Z}_{\text{Detached}}], h_{t-1})$
         \STATE $\ln B_{t}^{\text{Net}} \leftarrow \ln B_{t-1}^{\text{Net}} + \Delta $ 
       \ENDFOR

      \STATE Sample action $a \sim B_{Tg}^{\text{Net}}$ \hfill \COMMENT{Action taken at $t=T$}
      \STATE $\delta \leftarrow \mathbb{I}(a = r_g)$ \hfill \COMMENT{Sparse reward signal}
      \STATE 
     \STATE {\bfseries $\Phi$ Backward Pass:}
      \STATE $\mathcal{L}_{Clss} \leftarrow 0$
      \FOR{$t=1$ {\bfseries to} $T$}
      \STATE $B_{tga}^{\text{Net}} \leftarrow \text{soft clip}(B_{tga}^{\text{Net}}, \epsilon_{clip})$ \hfill \COMMENT{Soft clip}
      \STATE $\text{Reg} \leftarrow  -B_{tga}^{\text{Net}}\ln B_{tga}^{\text{Net}}$ \hfill \COMMENT{Regularization}
      \STATE $\text{PG} \leftarrow  \delta \ln B_{tga}^{\text{Net}} + (1-\delta)\ln(1-B_{tga}^{\text{Net}})$ \hfill \COMMENT{Policy Gradient}
       \STATE $\mathcal{L}_{Clss} \leftarrow \mathcal{L}_{Clss} - \frac{1}{T}(\text{PG} + \beta\cdot \text{Reg})$
      \ENDFOR
      \STATE Update $\Phi,h_{\phi}$ to minimize $\mathcal{L}_{Clss}$ \hfill \COMMENT{Gradient flows through $\Phi$}
      
      \STATE
      \STATE {\bfseries $\Psi$ Forward Pass:}
      \STATE  $\hat{r}_g \leftarrow a $ 
      \STATE $\hat{r}_c  \sim B_{Tc}^{\text{Net}}$ \hfill \COMMENT{Sample context realization}
      \STATE $\hat{\mathbf{r}}_{conf} \leftarrow B_{T\hat{\mathbf{r}}}^{\text{Net}}$, with index $g$ set to $\delta$ \hfill \COMMENT{Classifier confidence}
      \STATE $\hat{P}_{\hat{Z}}(\mathbf{o} \mid \hat{\mathbf{r}}) \leftarrow \Psi([\hat{Z}, \hat{r}_g,\hat{r}_c, \hat{\mathbf{r}}_{conf}])$ 
      
        \STATE 
      \STATE {\bfseries $\Psi$ Backward Pass:}
        \STATE $\mathbf{OPE}^i \leftarrow \frac{1}{2}J(\hat{P}_{\hat{Z}}(\mathbf{o} \mid \hat{\mathbf{r}}) \,\|\, \tilde{P} (\mathbf{o}))$  \hfill  \COMMENT{Observation Prediction Error (soft-clipped probabilities)}
        \STATE $\mathbf{REG}^{i}_V \leftarrow \langle\text{MSE}(\| f_V(V^i)\|_2 , 1)\rangle_c$ \hfill  \COMMENT{Embedding Regularization}

      \STATE $\mathcal{L}_{Gen} \leftarrow \mathcal{L}^{\star}_{\text{Generator}}$ \hfill \COMMENT{Reward-weighted $\mathbf{OPE}$ plus $\mathbf{REG}$, see \autoref{meth:exp2}} 
      \STATE Update $\Psi, \hat{K}_{\mathcal{C}}, \hat{Q}_{\mathcal{C}}, f_K, f_Q$ to minimize $\mathcal{L}_{Gen}$ \hfill \COMMENT{Gradient flows through $\Psi$ to Embeddings}
   \ENDFOR
\end{algorithmic}
\end{algorithm}
%%%%%%%%%%%%%%%%%%%%%%%%%%%%%%%%%%%%%%%%%%%%%%%%%%%%%%%%%%%%%%%%%%%%%%

\begin{algorithm}[H]
\caption{Experiment 3: Offline Controller Training}\label{alg:Controller_training}\begin{algorithmic}
\STATE {\bfseries Input:} Generator $\Psi$, Preferred Observation $\Omega$ (fixed for the run), a batch of contexts with estimated interactions $\{\hat{Z}\}$ (the same batch in every episode)
\STATE {\bfseries Hyperparameters:} Training Episodes $N_{\text{ep}}$, Entropy bonus weight $\beta_{cntrl}$
\STATE {\bfseries Models:} Actor $\Pi$, Critic $V$
\STATE
\STATE {\bfseries Intrinsic Reward Function $\mathcal{R}(\mathbf{r}, \mathbf{\Omega}, \ell)$:}
\STATE $\ell_{\hat{\mathbf{z}}^i} \leftarrow \Psi(\hat{\mathbf{z}}^i, \mathbf{r})$ \hfill \COMMENT{Generator provides likelihood estimates}
\STATE $\text{PL}^i = \ln(\ell_{\hat{\mathbf{z}}^i})\Omega^i + \ln(1-\ell_{\hat{\mathbf{z}}^i})(1-\Omega^i)$ \hfill \COMMENT{Preference logit}
\STATE $\text{Reward} \leftarrow \exp\langle \text{PL}\rangle_i$
\RETURN $\text{Reward}$ 

\STATE
\FOR{$e=1$ to $N_{\text{ep}}$}
  \FOR{each context $\hat{Z}$ in the batch}
    \STATE $\textbf{PE}\leftarrow -\sum_{\mathbf{r}}\pi(\mathbf{r} \mid \hat{Z})\ln\pi(\mathbf{r} \mid \hat{Z})$ \hfill \COMMENT{Compute policy entropy}
    \STATE $\mathbf{r} \sim \pi(\mathbf{r} \mid \hat{Z})$ \hfill \COMMENT{Sample joint action}
    \STATE $A \leftarrow \mathcal{R}(\mathbf{r}, \mathbf{\Omega}, \ell)-\nu(\hat{Z})$ \hfill \COMMENT{Compute Advantage}
    \STATE $\mathcal{L}_{A2C} \leftarrow  A ^2 - A_{\text{Detached}}\ln\pi(\mathbf{r} \mid \hat{Z}) -  \beta_{cntrl} \cdot \textbf{PE}$
  \ENDFOR
\STATE Update $\Pi, V$ to minimize $\mathcal{L}_{A2C}$ averaged over the batch
\ENDFOR
\end{algorithmic}
\end{algorithm}

\section*{Statement on LLM usage}

Large language models (LLMs) were used to assist with improving the clarity and style of the manuscript text including this statement; additionally, LLMs aided in providing relevant articles. All conceptual contributions, experimental design, and analysis were conducted by the authors.

\end{document}